\documentclass{article} 
\usepackage[preprint]{colm2026_conference}

\usepackage{microtype}
\usepackage{hyperref}
\usepackage{url}
\usepackage{booktabs}
\usepackage{amsmath}
\usepackage{enumitem}
\usepackage{graphicx}
\usepackage{xcolor}
\usepackage{tcolorbox}
\tcbuselibrary{skins,breakable}
\usepackage{wrapfig}
\usepackage{longtable}
\usepackage{colortbl}
\definecolor{rhoA}{HTML}{DCEAF7}  
\definecolor{rhoB}{HTML}{B5D4F0}  
\definecolor{rhoC}{HTML}{8DBDE6}  
\definecolor{rhoD}{HTML}{6BA3DB}  
\definecolor{rhoE}{HTML}{4A87CC}  
\definecolor{rhoF}{HTML}{2D6CB5}  
\definecolor{diaggray}{HTML}{F0F0F0}

\newcommand{\cA}[1]{\cellcolor{rhoA}{#1}}
\newcommand{\cB}[1]{\cellcolor{rhoB}{#1}}
\newcommand{\cC}[1]{\cellcolor{rhoC}{#1}}
\newcommand{\cD}[1]{\cellcolor{rhoD}{#1}}
\newcommand{\cE}[1]{\cellcolor{rhoE}\textcolor{white}{#1}}
\newcommand{\cF}[1]{\cellcolor{rhoF}\textcolor{white}{#1}}
\newcommand{\diag}{\cellcolor{diaggray}---}

\usepackage{lineno}
\usepackage{color-edits}
\addauthor{gn}{magenta}

\definecolor{darkblue}{rgb}{0, 0, 0.5}
\hypersetup{colorlinks=true, citecolor=darkblue, linkcolor=darkblue, urlcolor=darkblue}

\newcommand{\numModels}{9}
\newcommand{\numFamilies}{4}

\title{What Do Language Models Learn and When? The Implicit Curriculum Hypothesis}

\author{
Emmy Liu$^{1}$, Kaiser Sun$^{2}$, Millicent Li$^{3}$, Isabelle Lee$^{4}$, Lindia Tjuatja$^{1}$,\\ \hspace{0.5mm} \textbf{Jen-tse Huang$^{2}$, Graham Neubig$^{1}$} \\
$^{1}$Language Technologies Institute, Carnegie Mellon University \\
$^{2}$Department of Computer Science, Data Science and AI Institute, \\ \hspace{0.7mm} Johns Hopkins University \\
$^{3}$Khoury College of Computer Science, Northeastern University \\
$^{4}$Department of Computer Science, University of Southern California \\
\vspace{0.3em}
\texttt{emmy@cmu.edu}
\vspace{-1em}
}
\begin{document}

\ifcolmsubmission
\linenumbers
\fi

\maketitle
\begin{abstract}


Large language models (LLMs) can perform remarkably complex tasks, yet the fine-grained details of how these capabilities emerge during pretraining remain poorly understood. Scaling laws on validation loss tell us how much a model improves with additional compute, but not what skills it acquires in which order. To remedy this, we propose the \emph{Implicit Curriculum Hypothesis}: pretraining follows a compositional and predictable curriculum across models and data mixtures. We test this by designing a suite of simple, composable tasks spanning retrieval, morphological transformations, coreference, logical reasoning, and mathematics. Using these tasks, we track emergence points across four model families spanning sizes from 410M--13B parameters. We find that \emph{emergence orderings} of when models reach fixed accuracy thresholds are strikingly consistent ($\rho = .81$ across 45 model pairs), and that composite tasks most often emerge after their component tasks. Furthermore, we find that this structure is encoded in model representations: tasks with similar function vector representations also tend to follow similar trajectories in training. By using the space of representations derived from our task set, we can effectively predict the training trajectories of simple held-out compositional tasks throughout the course of pretraining ($R^2 = .68$--$.84$ across models) without previously evaluating them. Together, these results suggest that pretraining is more structured than loss curves reveal: skills emerge in a compositional order that is consistent across models and readable from their internals.\footnote{Data and code available at \url{https://github.com/KaiserWhoLearns/ElementalTask}}

\end{abstract}

\section{Introduction}


Large language models (LLMs) exhibit predictable improvements in performance with scale, a phenomenon characterized by well-established scaling laws \citep{hoffmann2022an,gadre2025language,muennighoff2023scaling}.
These scaling laws tell us how much models are expected to improve in predicting the next token on the pretraining distribution given additional compute, but not what skills the model acquires, or when during pretraining it acquires them specifically.
In practice, training runs may cost millions of dollars, yet are primarily monitored through aggregate cross-entropy loss, or through evaluating at intervals on downstream benchmarks such as MMLU \citep{mmlu}.
However, neither approach provides actionable diagnostic information. 
Cross-entropy loss decreases smoothly even as qualitatively different skills are acquired at sudden transition points \citep{kangaslahti2025hidden}.
Downstream benchmarks compose many prerequisite skills, making failures opaque: when GSM8k \citep{gsm8k} performance stalls, it is unclear whether the bottleneck is numerical fluency, multi-step planning, or natural language understanding.
For instance, scoring well on GSM8k may require numerical fluency, multi-step planning, as well as natural language understanding, making it difficult to diagnose which prerequisite skills are missing when performance stalls \citep{meister-cotterell-2021-language}.

A growing body of theoretical work suggests that neural networks learn functions sequentially, acquiring simpler patterns before more complex ones \citep{lee2025distinctcomputationsemergecompositional, zhang2026saddletosaddledynamicsexplainssimplicity}.
Recent work has built on these insights, hypothesizing that complex behaviors and scaling laws themselves emerge from the combination of more elementary sub-tasks that serve as fundamental building blocks \citep{khandelwal2025languagemodelscomposefunctions} or \textit{quanta} \citep{michaud2023the}.
However, much of this theoretical work has focused on simplified modeling settings, leaving open questions about how these insights translate to large-scale language model pretraining \citep{srivastava2023beyond}.
Prior empirical work has shown that certain knowledge categories (e.g., syntactic vs. factual) are acquired at different rates \citep{liu-etal-2021-probing-across}, and that grammatical phenomena are learned in a consistent order across architectures \citep{friedman-etal-2022-finding}. 
However, these studies have not examined whether the ordering reflects \textit{compositional} dependencies between skills, nor whether it is legible in the model's internal representations.

Based on this, we propose the \textbf{Implicit Curriculum Hypothesis}: \textbf{during pretraining, skills emerge in a stable compositional order that is consistent across models}.
This is a stronger claim than the quanta hypothesis alone.
It predicts not only that simple precedes complex, but also that the specific ordering is reproducible across models and reflects compositional dependencies between skills.
To test the Implicit Curriculum Hypothesis, we design a suite of simple tasks that probe a wide range of skills.
We track emergence across \numModels{} models from \numFamilies{} families (410M-13B parameters) and find:

\begin{enumerate}[leftmargin=*,itemsep=2pt,topsep=2pt]
\item \textbf{The emergence ordering is consistent across model families.}
Spearman correlations between emergence orderings range from $\rho = .64$ to $.93$ (mean $.81$) across all 45 model pairs, including cross-family comparisons. Copying is the first skill to emerge, followed by many simple string operations, fact extraction and coreference, then logic operations, simple world knowledge, then multistep arithmetic and more complex reasoning tasks.
Composite tasks emerge after their elemental prerequisites.
However, this consistency holds only when emergence is defined by fixed accuracy thresholds, not relative ones.

\item \textbf{The ordering is legible in model representations.}
Tasks whose internal representations are nearby in the model's residual stream, measured via function vectors \citep{todd2024function}, follow similar learning trajectories.
This proximity is sufficient to predict the full training trajectory of held-out composite tasks (mean $R^2$ of $.68$--$.84$ across models, with per-task $R^2$ exceeding $.95$) without ever evaluating them during training.
\end{enumerate}

\noindent Together, these results suggest that pretraining is more structured than loss curves reveal: skills emerge in an order that is consistent across models, respects compositional dependencies, and is readable from model internals. 

\begin{figure*}
    \centering
    \includegraphics[width=0.8\linewidth]{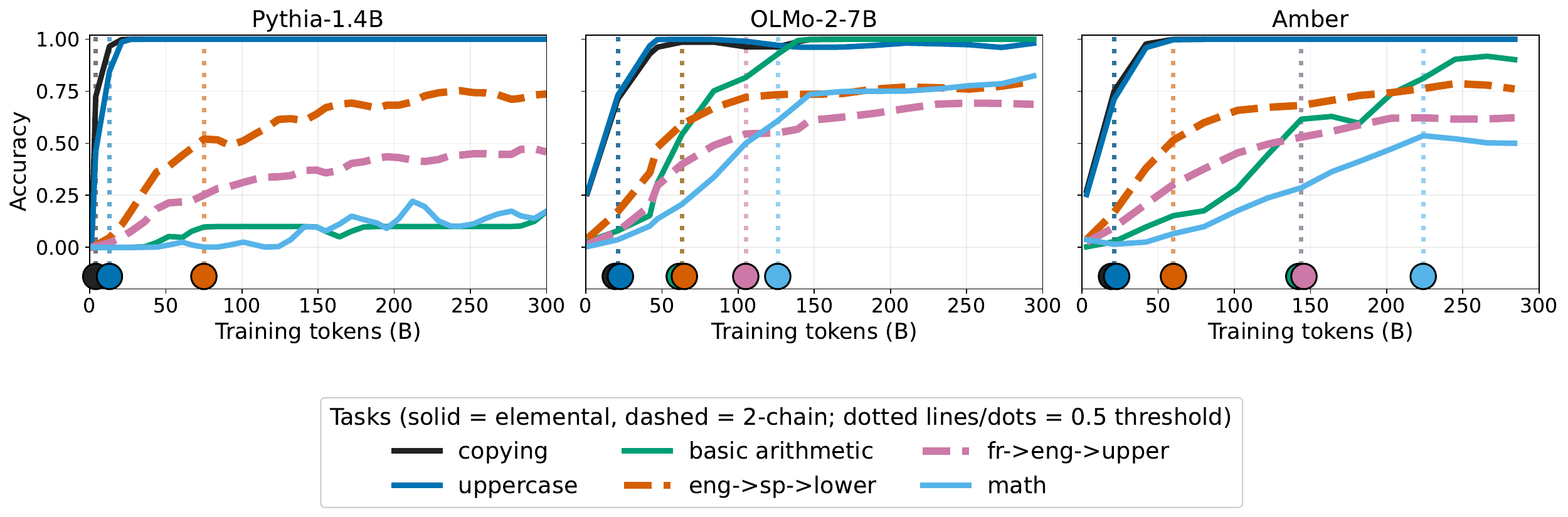}
    \caption{Emergence order across model families and sizes, smoothed with a Gaussian kernel ($\sigma=1.0$). Dots represent the point at which the model reaches a fixed 50\% accuracy threshold. While the absolute emergence time varies across models, the ordering shows regularity.}
    \label{fig:teaser_figure}
    \vspace{-3em}
\end{figure*}

\section{Preliminaries}

\subsection{Background}

We provide a summary of the work that we directly build on in this paper.
Further related work can be found in \autoref{appendix:related_work_extended}.

\paragraph{Scaling Laws}
Scaling laws characterize the relationship between a model's held-out validation loss $L$ and the compute budget allocated to training, typically decomposed into model size $N$ and data size $D$. These relationships are well-approximated by power laws of the form $L(N, D) \propto N^{\alpha} + D^{-\beta} + L_\infty$ \citep{kaplan2020scaling,hoffmann2022an}, and hold across many orders of magnitude. However, this aggregate loss curve does not directly correlate with downstream performance \citep{lourie2025scalinglawsunreliabledownstream, isik2026scalinglawsdownstreamtask, liu2026notjustscalinglawsbetterunderstanding}, and it is not clear what the model is learning as loss decreases.

\textbf{Quantization Hypothesis} \cite{michaud2023the} offers a hypothesis that these smooth scaling curves arise from the learning of discrete skills, termed \textit{quanta}.
Under this framework, a model acquires these quanta in an order optimized to reduce total loss, hiding the discrete transitions that correspond to the model learning.
One practical quandary is that quanta require a post-hoc discovery method, the results of which often do not correspond to interpretable skills \citep{michaud2023the}. 
While compelling, the Quantization Hypothesis typically treats these skills as independent, additive contributions, leaving the structural dependencies and compositional nature of these skills largely unexplored.

\paragraph{Simplicity Bias} 
Secondly, works show that neural networks trained with gradient-descent-based methods tend to exhibit \textbf{simplicity bias}, a tendency to learn simpler functions before more complex ones \citep{saxe2014exactsolutionsnonlineardynamics,nakkiran2019sgdneuralnetworkslearns, shah2020pitfallssimplicitybiasneural}. 
In the context of language modeling, this is reflected in models learning lower-order n-grams before higher-order ones \citep{michaelov2025languagemodelbehavioralphases}.
However, we note that the notions of simplicity are often underspecified, and moreover, these function-level notions of complexity cannot be used to quantify task complexity. 

\paragraph{Compositional Skill Structure}
A recent line of work has investigated whether skills acquired by language models follow explicit dependency structures. \citet{chen2023skillitdatadrivenskillsframework} represent skills as directed acyclic graphs (DAGs), where an edge from skill $A$ to skill $B$ indicates that training on data associated with $A$ reduces the amount of data needed to learn $B$.  Such dependency graphs can be used to design curricula for target skills.
While their work focuses on post-training and defines skills by data clusters, a natural question is whether we can also characterize the dependency structure of general web-data-based pretraining. Theoretically, \citet{arora2023theoryemergencecomplexskills} also provide a framework relating cross-entropy loss to competence on individual sets of skills, showing that a decrease in loss implies simultaneous improvement in both individual skills and their $k$-tuples. 

\subsection{The Implicit Curriculum Hypothesis}

The four threads of work above establish that (1) capabilities are discrete and unlock progressively, (2) simpler functions are learned before more complex ones, and (3) skills may have a dependency structure. However, they leave open whether these threads combine in practice: does large-scale pretraining follow a structured, compositional ordering of skill acquisition that is consistent across models? 

\begin{tcolorbox}[
  enhanced,
  breakable,
  colback=gray!5,
  colframe=gray!50,
  title={The Implicit Curriculum Hypothesis}
]
Let $\mathcal{T}$ be a set of tasks equipped with a \textit{known} dependency 
relation $\prec$, where $\tau_i \prec \tau_j$ indicates that task $\tau_j$ was 
constructed to compositionally depend on $\tau_i$. Let 
$P(\tau) = \{\tau' \in \mathcal{T} : \tau' \prec \tau\}$ denote the 
prerequisite set of task $\tau$ under this construction. We emphasize that 
$\prec$ reflects our \textit{design-level} task structure, not a claim about 
the model's internal primitives or the existence of any particular dependency 
in the model. Let $t^*_\tau(m)$ denote the emergence time of task $\tau$ for 
model $m$, defined as the first training step at which performance exceeds a 
threshold $\theta \in \mathbb{R}$. We hypothesize that this design-level 
structure is \textit{reflected empirically} in model training dynamics:

\begin{enumerate}[label=\textbf{H\arabic*.}]
    \item \textbf{Compositional ordering.} Tasks emerge no later than the 
    tasks constructed to depend on them:
    $$\forall\, \tau_j \in \mathcal{T},\; \forall\, \tau_i \in P(\tau_j): 
    \quad t^*_{\tau_i}(m) \leq t^*_{\tau_j}(m)$$

    \item \textbf{Cross-model stability.} The emergence ordering induces a 
    partial order $\preceq_{\mathcal{T}}$ over tasks that is consistent across 
    models: for models $m_1, m_2$, the rank correlation between their emergence orderings is significantly higher than chance.

    \item \textbf{Representational alignment.} Tasks with similar internal 
    representations exhibit similar learning trajectories. That is, for tasks 
    $\tau_i, \tau_j$ with representation vectors $v_i, v_j$, $\exists\, 
    \epsilon \in \mathbb{R}^{+}$:
    $$\mathrm{Sim}(v_i, v_j) \text{ high} \;\implies\; 
    d\!\left(a_{\tau_i}(\cdot),\; a_{\tau_j}(\cdot)\right) < \epsilon$$
    where $a_\tau(\cdot)$ is the learning trajectory of task $\tau$ and $d$ 
    is a distance over trajectories. This further implies that trajectories 
    of unseen tasks can be predicted from representational geometry alone, 
    without evaluating them during training.
\end{enumerate}
\end{tcolorbox}

\section{Methodology}

\subsection{Models and Checkpoints}

In order to test our hypotheses, we focus on examining open-weight models with publicly-released intermediate pre-training checkpoints. 
Because our hypotheses are largely about timing and emergence order, it was also important to select models with relatively dense intermediate checkpoints and larger sizes. 
The selected models are:

\begin{itemize}
    \item \textbf{OLMo-2} \citep{olmo20242}: 1B, 7B, and 13B
    parameter models, providing a within-family scale comparison
    across an order of magnitude.
    \item \textbf{OLMo-3} \citep{olmo3}: 7B, offering comparison with a newer generation compared to OLMo-2.
    \item \textbf{LLM360} \citep{liu2023llm360}: Crystal (7B)
    and Amber (7B), trained on very different data mixtures
    (code-oriented and natural-language-oriented, respectively),
    allowing us to study the effect of data composition within
    the same model family.
    \item \textbf{Pythia} \citep{biderman2023pythia}: 410M,
    1.4B, and 12B, offering a comparison with an earlier model generation
    trained on different data. We selected sizes spanning the
    full range of the suite; models below 410M were excluded
    due to poor performance.
\end{itemize}

In order to keep checkpoint sampling consistent across families, we focused on up to the first 1T tokens of training for each model and sampled approximately 20 checkpoints for each model within this range, giving a granularity of roughly every 20B tokens. We hypothesized that this would capture the period for which most relevant simple skills emerge for the tasks we study, while the granularity would be sufficient to resolve ordering differences.


\subsection{Task Design}
\label{sec:tasks}
We design tasks with intuitive compositional relationships, diverse operation types, and unambiguous outputs, while keeping them simple enough for models as small as 1B parameters to eventually solve via in-context learning. We therefore evaluate all 91 elemental and composite tasks using exact-match accuracy; the full list is given in \autoref{appendix:all_tasks}.

\begin{wraptable}{r}{0.60\textwidth}
\vspace{-0.8em}
\centering
\tiny
\setlength{\tabcolsep}{3pt}
\begin{tabular}{p{2.5cm}p{4.0cm}p{0.8cm}}
\toprule
\textbf{Task} & \textbf{Input} & \textbf{Out} \\
\midrule
\texttt{simple\_icl: present\_to\_gerund}
& run
& running \\
\addlinespace[0.3em]

\texttt{fact\_extraction: extract\_number}
& Passage: ``John gave 5 apples to Mary on Tuesday.'' How many apples?
& 5 \\
\addlinespace[0.3em]

\texttt{logical\_ops: conditional}
& If it rains, the ground gets wet. It rains. Is the ground wet?
& yes \\
\midrule
\multicolumn{3}{l}{\textit{Compositional tasks}} \\
\midrule

\texttt{comp: gerund\_upper}
& run
& RUNNING \\
\addlinespace[0.3em]

\texttt{comp: translate\_sp\_eng\_reverse}
& hola
& olleh \\
\addlinespace[0.3em]

\texttt{comp: extract\_verify}
& Passage: Nora gave 3 apples to Ben. Ben gave 1 to Li. Claim: Ben received apples first. Does the claim follow?
& True \\
\bottomrule
\end{tabular}
\caption{Example simple and compositional tasks. Compositional tasks require chaining multiple primitive skills.}
\label{tab:task-examples}
\vspace{-1em}
\end{wraptable}

\paragraph{Simple tasks.} We define a set of simple tasks spanning string manipulation (e.g., copy, uppercase, first letter), morphological transformation (e.g., singular to plural, present to gerund), knowledge retrieval (country to capital, country to currency), and translation (e.g., \texttt{en-fr}, \texttt{en-sp}). These were selected to cover distinct operation
types while remaining simple enough to be plausibly atomic. Notably, several of these operations have also been investigated in the interpretability literature \citep{olsson2022incontextlearninginductionheads, hendel-etal-2023-context, todd2024function, todd2026incontextalgebra}.  We do not claim that these are the true minimal units of model
computation, but they serve as a diverse set of operations from which we can construct composites with known structure. In total, we create 53 simple tasks.

\paragraph{Composite tasks: synthetic chains.} We construct composite tasks by chaining elemental operations in sequence. For example, \texttt{gerund\_upper} applies the gerund transformation followed by uppercasing (write $\to$ WRITING). This mechanical construction guarantees that the compositional prerequisites are known exactly, yielding 38 composite
tasks. The inclusion of translation-based composites (e.g., \texttt{translate\_eng\_fr\_upper\_reverse}) additionally tests whether knowledge-dependent elementals compose in the same way as rule-based ones.

\subsection{Measuring Emergence}

Prior work has proposed several notions of emergence, including scale-based definitions and parametric fits to learning curves \citep{wei2022emergentabilitieslargelanguage,snell2024predictingemergentcapabilitiesfinetuning}. For our purposes, however, the key quantity is not the sharpness of emergence but the \emph{relative ordering} of when tasks become feasible. Because many trajectories are noisy or irregular, we use simple threshold-based definitions. We consider two variants:

\paragraph{Absolute threshold.}
We define emergence time $t^*_\tau(m)$ as the first checkpoint at which task $\tau$ exceeds a fixed accuracy threshold $\theta_{\text{abs}}$.

\paragraph{Relative threshold.}
We alternatively define emergence time as the first checkpoint at which performance reaches a fraction $\alpha$ of the model's best performance on that task.

\subsection{Measuring Representational Similarity}

To operationalize representational alignment, we require a per-task representation that captures the computation the model performs in order to do the task. Following the methodology from \citet{todd2024function}, we extract task representations (function vectors) from the models.
\paragraph{Extraction.} Let a transformer have $L$ blocks and hidden dimension $d$. For block $\ell$, let
\[
h^{\mathrm{attn}}_{\ell}
=
h_{\ell-1}
+
\mathrm{Attn}\!\left(\mathrm{LN}(h_{\ell-1})\right)
\]
denote the post-attention hidden state, and let
\[
h_{\ell}
=
h^{\mathrm{attn}}_{\ell}
+
\mathrm{MLP}\!\left(\mathrm{LN}(h^{\mathrm{attn}}_{\ell})\right)
\]
denote the block-output hidden state. For each task $\tau$, we construct a set of ICL prompts, perform a forward pass for each prompt, and extract activations at the last non-pad token position $t_{\mathrm{last}}$ (i.e., the position from which the model begins generating its answer). We retain only prompts on which the model produces the correct answer, ensuring that the extracted representation reflects successful task execution. We consider two extraction methods, and for each model use the one that performs best (see \autoref{appendix:fv_hyperparams}).

\textit{Head-based extraction.} We use causal indirect effect (CIE) analysis to identify a sparse set of attention heads $\mathcal{H} \subseteq [H] \times [L]$ with the strongest causal effects on task performance. The function vector is then the average of these heads' outputs across correctly answered prompts:
$$
v_\tau^{\mathcal{H}} = \frac{1}{|\mathcal{D}_\tau^+|}
\sum_{x_i \in \mathcal{D}_\tau^+} \sum_{(h, j) \in \mathcal{H}} a_h^{j}(x_i),
$$
where $a_h^{j}(x_i)$ is the output of attention head $h$ in block $j$, evaluated at position $t_{\mathrm{last}}$, and $\mathcal{D}_\tau^+$ denotes the set of correctly answered prompts. We additionally constrain all selected heads to come from the same block.

\textit{Hidden-state extraction.} Alternatively, we extract the block-output hidden state at block $\ell$ and position $t_{\mathrm{last}}$:
$$
v_\tau^{\ell} = \frac{1}{|\mathcal{D}_\tau^+|}
\sum_{x_i \in \mathcal{D}_\tau^+}
h_{\ell, t_{\mathrm{last}}}(x_i),
$$
where $h_{\ell, t_{\mathrm{last}}}(x_i) \in \mathbb{R}^d$ is the post-MLP hidden state at block $\ell$ and position $t_{\mathrm{last}}$.

\paragraph{Task similarity} We measure similarity between tasks via cosine similarity between their task representations. The hypothesis predicts that tasks with higher representational similarity exhibit more similar learning trajectories.

\subsection{Evaluation Protocol}

We evaluate the Implicit Curriculum Hypothesis through two
complementary analyses, corresponding to the behavioral claims
(H1, H2) and the representational claim (H3).

\paragraph{Testing compositional ordering (H1).} 
For each composite task $c$ with a known set of prerequisite tasks $P(c)$, we check whether all prerequisites emerge no later than the composite:
$$\forall\, \tau \in P(c): \quad t^*_\tau(m) \leq t^*_c(m)$$
We report the violation rate: the fraction of (composite, prerequisite, model) triples for which this ordering is violated across all models. For synthetic chain composites, $P(c)$ is known by construction.

\paragraph{Testing cross-model stability (H2).} For each pair of models $(m_1, m_2)$, we compute the Spearman rank correlation between their emergence orderings $\sigma_{m_1}$, $\sigma_{m_2}$ over the full task set. We report correlations separately for the absolute and relative threshold definitions. For tasks that remain unemerged by the end of training, we bin them into one bucket at the end.\footnote{i.e. their emergence time is considered to be 1001B tokens for a 1T training run}


\paragraph{Leave-one-out prediction of composite trajectories (H3).}
We operationalize H3 through a leave-one-out (LOO) protocol over composite tasks. For a held-out composite task $c$, we predict its learning trajectory from the trajectories of its nearest neighbors in FV space.

Before prediction, we interpolate basis task trajectories onto the held-out task's token grid, apply Gaussian smoothing ($\sigma = 1.0$), and also discard tasks with near-zero trajectory variance.\footnote{in practice, this ended up being compositions of the \texttt{reverse} task as they were usually 0 throughout training.}

We extract unit-normalized residual stream representations for all tasks at the selected layer and compute pairwise similarities using an RBF kernel:
$$K(v_i, v_j) = \exp\!\Big(\!-\frac{\|v_i - v_j\|^2}
{2\sigma_k^2}\Big)$$
We use kernel ridge regression to learn a predictor for the held-out task performance: Let $S$ denote the set of training tasks (excluding $c$), let $K_S \in \mathbb{R}^{|S| \times |S|}$ be the kernel matrix with entries
$$
(K_S)_{ij} = K(v_{\tau_i}, v_{\tau_j}),
$$
and let
$$
k_c = \big[K(v_c, v_{\tau_j})\big]_{j \in S}
$$
be the vector of similarities between the held-out task and the training tasks. For each training step $t$, we form the vector of training trajectory values
$$
y_t = \big[a_{\tau_j}(t)\big]_{j \in S}.
$$
Kernel ridge regression solves
$$
\alpha_t = (K_S + \lambda I)^{-1} y_t
$$
and predicts the held-out trajectory as
$$
\hat{a}_c(t) = k_c^\top \alpha_t.
$$

We evaluate prediction quality via per-task Pearson $r^2$ and MAE against smoothed ground-truth trajectories, and report both per-task results and means across all held-out composites.
To test the composition bottleneck, we compare two conditions for the function vector space:
\begin{enumerate}
    \item \textbf{All tasks:} the basis includes both simple
    and composite tasks excluding the held-out
    target.
    \item \textbf{Simple tasks only:} the basis is restricted to
    non-composite tasks only.
\end{enumerate}
If prediction quality degrades substantially under the elementals-only basis, this indicates that composite trajectories share structure with one another that is not captured by their elemental components alone, indicating a composition bottleneck.

\section{Emergence Order Results}


\begin{figure}
    \centering
    \vspace{-1em}
    \includegraphics[width=0.7\textwidth]{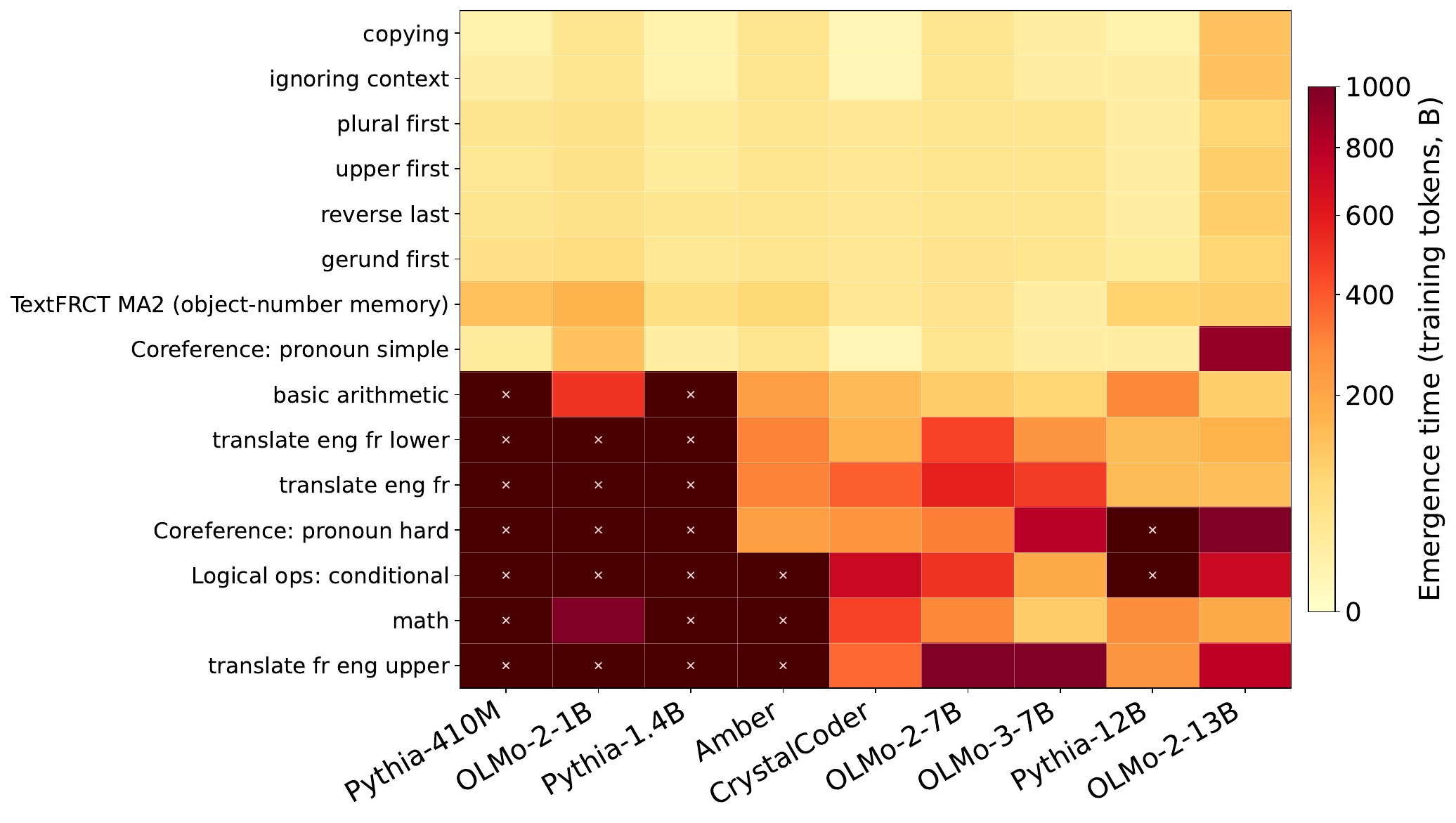}
    \caption{Emergence order heatmap of selected tasks across models (absolute threshold = 0.8). Tasks sorted by consensus emergence order. Consistent color gradients across columns indicate stable ordering.}
    \label{fig:heatmap}
    \vspace{-1em}
\end{figure}

We first test H1 and H2 by examining whether the emergence order of tasks is consistent across models and whether composites emerge after their constituent components. \autoref{fig:heatmap} shows the emergence times of all
tasks across models, sorted vertically by consensus emergence order.
From inspection, it is clear that tasks that emerge early in one model tend to do so across all models. Copying and simple coreference resolution emerge early across all models, in line with previous work \cite{yin2025which}. 
These are followed by simple ICL tasks such as uppercasing and lowercasing, then morphological transformations, followed by knowledge-dependent tasks such as translation, and finally
a long tail of more difficult or compositional tasks. Furthermore, we examine whether compositional tasks arise after their components. This is generally the case: among all compositional tasks, 54/76 emerged no earlier than their ``parent'' tasks. 
However, we also observe a number of inversions, where the composite task emerged earlier than one (19 weak inversions) or both (3 strong inversions) parents. Notably, all three strong inversions involve the \texttt{first\_letter} component task. 

\autoref{tab:spearman} quantifies the consistency of emergence order across models. Within the OLMo-2 family, Spearman rank correlations range from .72 to .93. 
Cross-family correlations are also high: Amber correlates with OLMo-2 models at 0.82–0.88, while correlations with older and smaller models (e.g., OLMo-2 vs. Pythia-410M) remain substantial, ranging from 0.64 to 0.84.
All correlations are highly significant and remain so after correction for multiple comparisons. 
Importantly, this consistency holds only under the absolute threshold definition of emergence. When using relative thresholds, cross-model correlations drop substantially (\autoref{appendix:emergence_orders_alternate}). 
We hypothesize that this discrepancy arises because relative thresholds depend on each model’s maximum performance: a weak model may reach a relative threshold early despite lacking meaningful task competence, while a stronger model may never satisfy the same criterion.
In contrast, our absolute thresholds are set above chance for all tasks, effectively capturing the point at which the underlying computation becomes functional.
This plausibly corresponds to the formation of a task-relevant circuit. 
In this view, the relative consistency of the absolute order suggests that what is shared across models is \textit{the order in which computations become feasible} under standard pretraining, even when trained on differing data distributions. 

\begin{table}[t]
\centering
\caption{Spearman rank correlation ($\rho$) of emergence orderings between model pairs (absolute threshold = 80\%). All 45 correlations are significant ($p<10^{-7}$).}
\label{tab:spearman}
\scriptsize
\setlength{\tabcolsep}{3.5pt}
\renewcommand{\arraystretch}{1.02}
\begin{tabular}{lcccccccccc}
\toprule
& \multicolumn{3}{c}{OLMo2} & OLMo3 & \multicolumn{2}{c}{LLM360} & \multicolumn{4}{c}{Pythia} \\
\cmidrule(lr){2-4}\cmidrule(lr){5-5}\cmidrule(lr){6-7}\cmidrule(lr){8-11}
& 1B & 7B & 13B & 7B & Amber & Crystal & 410M & 1.4B & 2.8B & 12B \\
\midrule
O2-1B   & \diag & \cD{.80} & \cB{.72} & \cC{.75} & \cE{.88} & \cC{.76} & \cD{.84} & \cF{.92} & \cC{.78} & \cD{.84} \\
O2-7B   &       & \diag    & \cF{.93} & \cF{.93} & \cE{.88} & \cF{.92} & \cB{.71} & \cB{.72} & \cC{.77} & \cD{.84} \\
O2-13B  &       &          & \diag    & \cF{.93} & \cD{.82} & \cE{.89} & \cA{.64} & \cA{.66} & \cC{.76} & \cD{.83} \\
O3-7B   &       &          &          & \diag    & \cD{.83} & \cF{.92} & \cB{.70} & \cB{.72} & \cC{.79} & \cE{.85} \\
Amber   &       &          &          &          & \diag    & \cE{.87} & \cC{.77} & \cD{.84} & \cE{.87} & \cF{.90} \\
Crystal &       &          &          &          &          & \diag    & \cB{.70} & \cC{.75} & \cD{.83} & \cE{.87} \\
P-410M  &       &          &          &          &          &          & \diag    & \cE{.88} & \cB{.72} & \cC{.79} \\
P-1.4B  &       &          &          &          &          &          &          & \diag    & \cC{.77} & \cD{.82} \\
P-2.8B  &       &          &          &          &          &          &          &          & \diag    & \cF{.91} \\
P-12B   &       &          &          &          &          &          &          &          &          & \diag \\
\bottomrule
\end{tabular}

\vspace{0.2em}

{\footnotesize
\colorbox{rhoA}{\strut\hspace{0.45em}} .64\;
\colorbox{rhoB}{\strut\hspace{0.45em}} .70\;
\colorbox{rhoC}{\strut\hspace{0.45em}} .75\;
\colorbox{rhoD}{\strut\hspace{0.45em}} .80\;
\colorbox{rhoE}{\strut\hspace{0.45em}} .85\;
\colorbox{rhoF}{\strut\hspace{0.45em}} .90+
\vspace{-2em}
}

\end{table}

\section{Representational Similarity and Prediction Results}



Having established that skill acquisition during pretraining is both structured and consistent (H1, H2), we next ask whether this structure is reflected in the model's internal representations (H3). Namely, if two tasks have similar function vectors, do they exhibit similar learning trajectories in pretraining? Rather than testing correlations in isolation, we consider a stronger version: can the learning trajectory of a held-out composite task be predicted solely from its representational similarity to other tasks, without further evaluation during training?

\begin{table}[t]
\centering
\caption{Leave-one-out prediction of held-out composite task trajectories (26 tasks). \textit{All tasks} includes simple and composite tasks. \textit{Sim.\ only} includes only simple tasks. Restricting to elementals degrades MAE for every model (mean $\Delta$MAE $= +.135$), indicating a composition bottleneck.}
\label{tab:loo}
\scriptsize
\begin{tabular}{l cc cc}
\toprule
 & \multicolumn{2}{c}{\textbf{All tasks}} & \multicolumn{2}{c}{\textbf{Sim.\ only}} \\
\cmidrule(lr){2-3} \cmidrule(lr){4-5}
\textbf{Model} & $R^2$ & MAE & $R^2$ & MAE \\
\midrule
Pythia-410M & .681 & .195 & .717 & .301 \\
OLMo2-1B    & .723 & .070 & .602 & .289 \\
Pythia-1.4B & .778 & .086 & .755 & .193 \\
\addlinespace[3pt]
Amber (7B)  & .751 & .082 & .725 & .205 \\
Crystal (7B)& .676 & .133 & .568 & .315 \\
OLMo2-7B    & .767 & .068 & .693 & .208 \\
OLMo3-7B    & .692 & .079 & .491 & .215 \\
\addlinespace[3pt]
Pythia-12B  & .812 & .136 & .789 & .194 \\
OLMo2-13B   & .838 & .099 & .860 & .242 \\
\bottomrule
\end{tabular}
\end{table}

\begin{figure}
    \centering
    \includegraphics[width=0.8\linewidth]{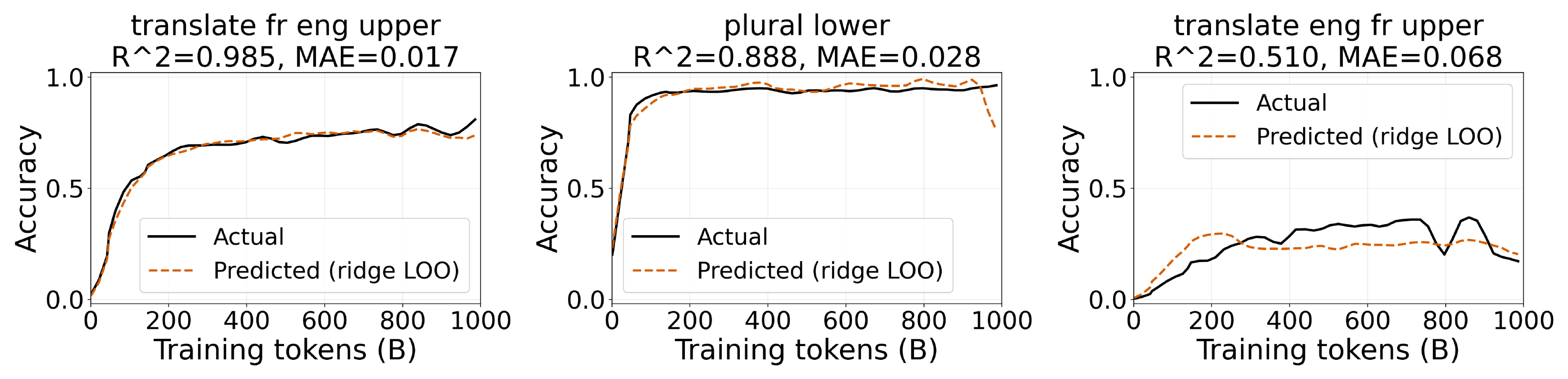}
    \caption{Example composite task predictions for OLMo2-7B between 0-1T tokens.}
    \label{fig:loo_examples}
\end{figure}

\autoref{tab:loo} reports leave-one-out prediction results for composite task trajectories using kernel ridge regression in function vector space. 
When the basis includes all other tasks (elemental and composite), prediction quality is strong: $R^2$ ranges from .67 (Crystal) to .838 (OLMo2-13B), with MAE between .068 and .195 on a 0-1 accuracy scale.
These results provide strong evidence that representational geometry is closely linked to learning dynamics, supporting H3.
As a case study of specific predicted trajectories, \autoref{fig:loo_examples} shows representative predicted trajectories compared to ground truth trajectories for OLMo2-7B from 0-1T tokens. For tasks such as \texttt{fr\_eng\_upper} ($R^2 = .99$, MAE $= .017$) and \texttt{plural\_lower} ($R^2 = .89$, MAE $= .028$), the predicted curve closely tracks the actual trajectory, capturing both the onset of emergence and the subsequent rate of improvement. However, predictions are weaker for tasks such as \texttt{eng\_fr\_upper} ($R^2 = .51$, MAE $= .068$), where the held-out task's trajectory is less well approximated by its nearest neighbors in representation space. 
Full prediction results can be found in \autoref{appendix:all_held_out_preds}.

\section{Conclusion}

In this paper, we unify several threads of discussion on the emergence of LM capabilities during pretraining in the Implicit Curriculum Hypothesis -- that skills acquired during pretraining emerge in a stable order driven by concrete compositions. We test this hypothesis empirically across several model families and with models spanning 410M-13B parameters. Our empirical findings support both the behavioural and representational aspects of the hypothesis: emergence orders of tasks under absolute thresholds are quite consistent across models, even across families and in models trained on different data. Furthermore, similarity in the function vector space predicts similarity of learning trajectories, such that it is possible to predict the trajectories of held-out compositional tasks from function vector similarity without evaluating them. This indicates that the developmental structure visible in
behavioral evaluations may also be legible in the model's internal
representations.

Our results open several avenues for further investigation. One practical application is \textit{pretraining monitoring} -- if emergence orders are stable and predictable, our task suite can serve as a basis for monitoring whether models are developing capabilities ahead of or behind schedule. Furthermore, understanding this task structure could also inform data mixture decisions. More broadly, we hope that the framework of studying pretraining as a structured developmental process will prove useful for understanding, predicting, and ultimately steering what language models learn.


\section*{Acknowledgments}
EL was supported by the National Sciences and Engineering Research Council of Canada (NSERC), [funding reference number 578085], as well as the SoftBank-ARM Fellowship. ML is supported by a NSF Graduate Research Fellowship. IL is supported by a Technical AI Safety Research Grant from Coefficient Giving via Berkeley Existential Risk Initiative.

This work used the Delta system at the National Center for Supercomputing Applications [award OAC 2005572] through allocation [CIS250578] from the Advanced Cyberinfrastructure Coordination Ecosystem: Services \& Support (ACCESS) program, which is supported by National Science Foundation grants \#2138259, \#2138286, \#2138307, \#2137603, and \#2138296.


\bibliography{reference, model}

@inproceedings{olmo2,
  title={2 OLMo 2 Furious},
  author={Walsh, Evan Pete and Soldaini, Luca and Groeneveld, Dirk and Lo, Kyle and Arora, Shane and Bhagia, Akshita and Gu, Yuling and Huang, Shengyi and Jordan, Matt and Lambert, Nathan and others},
  booktitle={Second Conference on Language Modeling},
  year={2025}
}

@article{olmo3,
  title={Olmo 3: Charting a path through the model flow to lead open-source AI},
  author={Olmo},
  journal={AI2 Blog Nov 30 2025},
  url={https://allenai.org/blog/olmo3},
  year={2025}
}

@article{olmo20242,
  title={2 OLMo 2 Furious},
  author={OLMo, Team and Walsh, Pete and Soldaini, Luca and Groeneveld, Dirk and Lo, Kyle and Arora, Shane and Bhagia, Akshita and Gu, Yuling and Huang, Shengyi and Jordan, Matt and others},
  journal={arXiv preprint arXiv:2501.00656},
  year={2024}
}

@inproceedings{biderman2023pythia,
  title={Pythia: A suite for analyzing large language models across training and scaling},
  author={Biderman, Stella and Schoelkopf, Hailey and Anthony, Quentin Gregory and Bradley, Herbie and O’Brien, Kyle and Hallahan, Eric and Khan, Mohammad Aflah and Purohit, Shivanshu and Prashanth, USVSN Sai and Raff, Edward and others},
  booktitle={International conference on machine learning},
  pages={2397--2430},
  year={2023},
  organization={PMLR}
}

@misc{liu2023llm360,
      title={LLM360: Towards Fully Transparent Open-Source LLMs}, 
      author={Zhengzhong Liu and Aurick Qiao and Willie Neiswanger and Hongyi Wang and Bowen Tan and Tianhua Tao and Junbo Li and Yuqi Wang and Suqi Sun and Omkar Pangarkar and Richard Fan and Yi Gu and Victor Miller and Yonghao Zhuang and Guowei He and Haonan Li and Fajri Koto and Liping Tang and Nikhil Ranjan and Zhiqiang Shen and Xuguang Ren and Roberto Iriondo and Cun Mu and Zhiting Hu and Mark Schulze and Preslav Nakov and Tim Baldwin and Eric P. Xing},
      year={2023},
      eprint={2312.06550},
      archivePrefix={arXiv},
      primaryClass={cs.CL}
}

@article{ge2025evolution,
  title={Evolution of Concepts in Language Model Pre-Training},
  author={Ge, Xuyang and Shu, Wentao and Wu, Jiaxing and Zhou, Yunhua and He, Zhengfu and Qiu, Xipeng},
  journal={arXiv preprint arXiv:2509.17196},
  year={2025}
}

@article{van2025polypythias,
  title={Polypythias: Stability and outliers across fifty language model pre-training runs},
  author={van der Wal, Oskar and Lesci, Pietro and Muller-Eberstein, Max and Saphra, Naomi and Schoelkopf, Hailey and Zuidema, Willem and Biderman, Stella},
  journal={arXiv preprint arXiv:2503.09543},
  year={2025}
}

@inproceedings{mishranext,
  title={From Next-Token to Mathematics: The Learning Dynamics of Mathematical Reasoning in Language Models},
  author={Mishra, Shubhra and Poesia, Gabriel and Goodman, Noah},
  booktitle={Second Conference on Language Modeling},
  year={2025}

}

@article{chen2023sudden,
  title={Sudden drops in the loss: Syntax acquisition, phase transitions, and simplicity bias in MLMs},
  author={Chen, Angelica and Shwartz-Ziv, Ravid and Cho, Kyunghyun and Leavitt, Matthew L and Saphra, Naomi},
  journal={arXiv preprint arXiv:2309.07311},
  year={2023}
}

@inproceedings{sun-dredze-2025-amuro,
    title = "Amuro {\&} Char: Analyzing the Relationship between Pre-Training and Fine-Tuning of Large Language Models",
    author = "Sun, Kaiser  and
      Dredze, Mark",
    editor = "Adlakha, Vaibhav  and
      Chronopoulou, Alexandra  and
      Li, Xiang Lorraine  and
      Majumder, Bodhisattwa Prasad  and
      Shi, Freda  and
      Vernikos, Giorgos",
    booktitle = "Proceedings of the 10th Workshop on Representation Learning for NLP (RepL4NLP-2025)",
    month = may,
    year = "2025",
    address = "Albuquerque, NM",
    publisher = "Association for Computational Linguistics",
    url = "https://aclanthology.org/2025.repl4nlp-1.11/",
    doi = "10.18653/v1/2025.repl4nlp-1.11",
    pages = "131--151",
    ISBN = "979-8-89176-245-9"
}

@article{kangaslahti2025hidden,
  title={Hidden breakthroughs in language model training},
  author={Kangaslahti, Sara and Rosenfeld, Elan and Saphra, Naomi},
  journal={arXiv preprint arXiv:2506.15872},
  year={2025}
}

@inproceedings{meister-cotterell-2021-language,
    title = "Language Model Evaluation Beyond Perplexity",
    author = "Meister, Clara  and
      Cotterell, Ryan",
    editor = "Zong, Chengqing  and
      Xia, Fei  and
      Li, Wenjie  and
      Navigli, Roberto",
    booktitle = "Proceedings of the 59th Annual Meeting of the Association for Computational Linguistics and the 11th International Joint Conference on Natural Language Processing (Volume 1: Long Papers)",
    month = aug,
    year = "2021",
    address = "Online",
    publisher = "Association for Computational Linguistics",
    url = "https://aclanthology.org/2021.acl-long.414/",
    doi = "10.18653/v1/2021.acl-long.414",
    pages = "5328--5339"
}

@inproceedings{
    michaud2023the,
    title={The Quantization Model of Neural Scaling},
    author={Eric J Michaud and Ziming Liu and Uzay Girit and Max Tegmark},
    booktitle={Thirty-seventh Conference on Neural Information Processing Systems},
    year={2023},
    url={https://openreview.net/forum?id=3tbTw2ga8K}
}

@inproceedings{
hoffmann2022an,
title={An empirical analysis of compute-optimal large language model training},
author={Jordan Hoffmann and Sebastian Borgeaud and Arthur Mensch and Elena Buchatskaya and Trevor Cai and Eliza Rutherford and Diego de las Casas and Lisa Anne Hendricks and Johannes Welbl and Aidan Clark and Tom Hennigan and Eric Noland and Katherine Millican and George van den Driessche and Bogdan Damoc and Aurelia Guy and Simon Osindero and Karen Simonyan and Erich Elsen and Oriol Vinyals and Jack William Rae and Laurent Sifre},
booktitle={Advances in Neural Information Processing Systems},
editor={Alice H. Oh and Alekh Agarwal and Danielle Belgrave and Kyunghyun Cho},
year={2022},
url={https://openreview.net/forum?id=iBBcRUlOAPR}
}

@inproceedings{
gadre2025language,
title={Language models scale reliably with over-training and on downstream tasks},
author={Samir Yitzhak Gadre and Georgios Smyrnis and Vaishaal Shankar and Suchin Gururangan and Mitchell Wortsman and Rulin Shao and Jean Mercat and Alex Fang and Jeffrey Li and Sedrick Keh and Rui Xin and Marianna Nezhurina and Igor Vasiljevic and Luca Soldaini and Jenia Jitsev and Alex Dimakis and Gabriel Ilharco and Pang Wei Koh and Shuran Song and Thomas Kollar and Yair Carmon and Achal Dave and Reinhard Heckel and Niklas Muennighoff and Ludwig Schmidt},
booktitle={The Thirteenth International Conference on Learning Representations},
year={2025},
url={https://openreview.net/forum?id=iZeQBqJamf}
}

@inproceedings{
muennighoff2023scaling,
title={Scaling Data-Constrained Language Models},
author={Niklas Muennighoff and Alexander M Rush and Boaz Barak and Teven Le Scao and Nouamane Tazi and Aleksandra Piktus and Sampo Pyysalo and Thomas Wolf and Colin Raffel},
booktitle={Thirty-seventh Conference on Neural Information Processing Systems},
year={2023},
url={https://openreview.net/forum?id=j5BuTrEj35}
}

@inproceedings{
todd2024function,
title={Function Vectors in Large Language Models},
author={Eric Todd and Millicent Li and Arnab Sen Sharma and Aaron Mueller and Byron C Wallace and David Bau},
booktitle={The Twelfth International Conference on Learning Representations},
year={2024},
url={https://openreview.net/forum?id=AwyxtyMwaG}
}

@inproceedings{hendel-etal-2023-context,
    title = "In-Context Learning Creates Task Vectors",
    author = "Hendel, Roee  and
      Geva, Mor  and
      Globerson, Amir",
    editor = "Bouamor, Houda  and
      Pino, Juan  and
      Bali, Kalika",
    booktitle = "Findings of the Association for Computational Linguistics: EMNLP 2023",
    month = dec,
    year = "2023",
    address = "Singapore",
    publisher = "Association for Computational Linguistics",
    url = "https://aclanthology.org/2023.findings-emnlp.624/",
    doi = "10.18653/v1/2023.findings-emnlp.624",
    pages = "9318--9333"
}

@misc{khandelwal2025languagemodelscomposefunctions,
      title={How Do Language Models Compose Functions?}, 
      author={Apoorv Khandelwal and Ellie Pavlick},
      year={2025},
      eprint={2510.01685},
      archivePrefix={arXiv},
      primaryClass={cs.CL},
      url={https://arxiv.org/abs/2510.01685}, 
}

@article{srivastava2023beyond,
  title={Beyond the Imitation Game: Quantifying and extrapolating the capabilities of language models},
  author={Srivastava, Aarohi and Rastogi, Abhinav and Rao, Abhishek and Shoeb, Abu Awal Md and Abid, Abubakar and Fisch, Adam and Brown, Adam R and Santoro, Adam and Gupta, Aditya and Garriga-Alonso, Adri{\`a} and others},
  journal={Transactions on Machine Learning Research},
  year={2023}
}

@misc{prasad2026featuresrewardsscalablesupervision,
      title={Features as Rewards: Scalable Supervision for Open-Ended Tasks via Interpretability}, 
      author={Aaditya Vikram Prasad and Connor Watts and Jack Merullo and Dhruvil Gala and Owen Lewis and Thomas McGrath and Ekdeep Singh Lubana},
      year={2026},
      eprint={2602.10067},
      archivePrefix={arXiv},
      primaryClass={cs.LG},
      url={https://arxiv.org/abs/2602.10067}, 
}

@inproceedings{friedman-etal-2022-finding,
    title = "Finding Dataset Shortcuts with Grammar Induction",
    author = "Friedman, Dan  and
      Wettig, Alexander  and
      Chen, Danqi",
    editor = "Goldberg, Yoav  and
      Kozareva, Zornitsa  and
      Zhang, Yue",
    booktitle = "Proceedings of the 2022 Conference on Empirical Methods in Natural Language Processing",
    month = dec,
    year = "2022",
    address = "Abu Dhabi, United Arab Emirates",
    publisher = "Association for Computational Linguistics",
    url = "https://aclanthology.org/2022.emnlp-main.293/",
    doi = "10.18653/v1/2022.emnlp-main.293",
    pages = "4345--4363"
}

@misc{arora2023theoryemergencecomplexskills,
      title={A Theory for Emergence of Complex Skills in Language Models}, 
      author={Sanjeev Arora and Anirudh Goyal},
      year={2023},
      eprint={2307.15936},
      archivePrefix={arXiv},
      primaryClass={cs.LG},
      url={https://arxiv.org/abs/2307.15936}, 
}

@misc{todd2026incontextalgebra,
      title={In-Context Algebra}, 
      author={Eric Todd and Jannik Brinkmann and Rohit Gandikota and David Bau},
      year={2026},
      eprint={2512.16902},
      archivePrefix={arXiv},
      primaryClass={cs.CL},
      url={https://arxiv.org/abs/2512.16902}, 
}

@misc{maimon2025iqtestllmsevaluation,
      title={IQ Test for LLMs: An Evaluation Framework for Uncovering Core Skills in LLMs}, 
      author={Aviya Maimon and Amir DN Cohen and Gal Vishne and Shauli Ravfogel and Reut Tsarfaty},
      year={2025},
      eprint={2507.20208},
      archivePrefix={arXiv},
      primaryClass={cs.CL},
      url={https://arxiv.org/abs/2507.20208}, 
}

@inproceedings{
    yu2024skillmix,
    title={{SKILL}-{MIX}: a Flexible and Expandable Family of Evaluations for {AI} Models},
    author={Dingli Yu and Simran Kaur and Arushi Gupta and Jonah Brown-Cohen and Anirudh Goyal and Sanjeev Arora},
    booktitle={The Twelfth International Conference on Learning Representations},
    year={2024},
    url={https://openreview.net/forum?id=Jf5gplvglq}
}

@misc{burnell2023revealingstructurelanguagemodel,
      title={Revealing the structure of language model capabilities}, 
      author={Ryan Burnell and Han Hao and Andrew R. A. Conway and Jose Hernandez Orallo},
      year={2023},
      eprint={2306.10062},
      archivePrefix={arXiv},
      primaryClass={cs.CL},
      url={https://arxiv.org/abs/2306.10062}, 
}

@inproceedings{
polo2025sloth,
title={Sloth: scaling laws for {LLM} skills to predict multi-benchmark performance across families},
author={Felipe Maia Polo and Seamus Somerstep and Leshem Choshen and Yuekai Sun and Mikhail Yurochkin},
booktitle={The Thirty-ninth Annual Conference on Neural Information Processing Systems},
year={2025},
url={https://openreview.net/forum?id=9GN5Jsa3lv}
}

@misc{mmlu,
      title={Measuring Massive Multitask Language Understanding}, 
      author={Dan Hendrycks and Collin Burns and Steven Basart and Andy Zou and Mantas Mazeika and Dawn Song and Jacob Steinhardt},
      year={2021},
      eprint={2009.03300},
      archivePrefix={arXiv},
      primaryClass={cs.CY},
      url={https://arxiv.org/abs/2009.03300}, 
}

@misc{gsm8k,
      title={Training Verifiers to Solve Math Word Problems}, 
      author={Karl Cobbe and Vineet Kosaraju and Mohammad Bavarian and Mark Chen and Heewoo Jun and Lukasz Kaiser and Matthias Plappert and Jerry Tworek and Jacob Hilton and Reiichiro Nakano and Christopher Hesse and John Schulman},
      year={2021},
      eprint={2110.14168},
      archivePrefix={arXiv},
      primaryClass={cs.LG},
      url={https://arxiv.org/abs/2110.14168}, 
}

@misc{nakkiran2019sgdneuralnetworkslearns,
      title={SGD on Neural Networks Learns Functions of Increasing Complexity}, 
      author={Preetum Nakkiran and Gal Kaplun and Dimitris Kalimeris and Tristan Yang and Benjamin L. Edelman and Fred Zhang and Boaz Barak},
      year={2019},
      eprint={1905.11604},
      archivePrefix={arXiv},
      primaryClass={cs.LG},
      url={https://arxiv.org/abs/1905.11604}, 
}

@misc{saxe2014exactsolutionsnonlineardynamics,
      title={Exact solutions to the nonlinear dynamics of learning in deep linear neural networks}, 
      author={Andrew M. Saxe and James L. McClelland and Surya Ganguli},
      year={2014},
      eprint={1312.6120},
      archivePrefix={arXiv},
      primaryClass={cs.NE},
      url={https://arxiv.org/abs/1312.6120}, 
}

@misc{shah2020pitfallssimplicitybiasneural,
      title={The Pitfalls of Simplicity Bias in Neural Networks}, 
      author={Harshay Shah and Kaustav Tamuly and Aditi Raghunathan and Prateek Jain and Praneeth Netrapalli},
      year={2020},
      eprint={2006.07710},
      archivePrefix={arXiv},
      primaryClass={cs.LG},
      url={https://arxiv.org/abs/2006.07710}, 
}

@misc{michaelov2025languagemodelbehavioralphases,
      title={Language Model Behavioral Phases are Consistent Across Architecture, Training Data, and Scale}, 
      author={James A. Michaelov and Roger P. Levy and Benjamin K. Bergen},
      year={2025},
      eprint={2510.24963},
      archivePrefix={arXiv},
      primaryClass={cs.CL},
      url={https://arxiv.org/abs/2510.24963}, 
}

@misc{kaplan2020scaling,
      title={Scaling Laws for Neural Language Models}, 
      author={Jared Kaplan and Sam McCandlish and Tom Henighan and Tom B. Brown and Benjamin Chess and Rewon Child and Scott Gray and Alec Radford and Jeffrey Wu and Dario Amodei},
      year={2020},
      eprint={2001.08361},
      archivePrefix={arXiv},
      primaryClass={cs.LG},
      url={https://arxiv.org/abs/2001.08361}, 
}

@misc{isik2026scalinglawsdownstreamtask,
      title={Scaling Laws for Downstream Task Performance of Large Language Models}, 
      author={Berivan Isik and Natalia Ponomareva and Hussein Hazimeh and Dimitris Paparas and Sergei Vassilvitskii and Sanmi Koyejo},
      year={2026},
      eprint={2402.04177},
      archivePrefix={arXiv},
      primaryClass={cs.CL},
      url={https://arxiv.org/abs/2402.04177}, 
}

@misc{liu2026notjustscalinglawsbetterunderstanding,
      title={Not-Just-Scaling Laws: Towards a Better Understanding of the Downstream Impact of Language Model Design Decisions}, 
      author={Emmy Liu and Amanda Bertsch and Lintang Sutawika and Lindia Tjuatja and Patrick Fernandes and Lara Marinov and Michael Chen and Shreya Singhal and Carolin Lawrence and Aditi Raghunathan and Kiril Gashteovski and Graham Neubig},
      year={2026},
      eprint={2503.03862},
      archivePrefix={arXiv},
      primaryClass={cs.CL},
      url={https://arxiv.org/abs/2503.03862}, 
}

@misc{lourie2025scalinglawsunreliabledownstream,
      title={Scaling Laws Are Unreliable for Downstream Tasks: A Reality Check}, 
      author={Nicholas Lourie and Michael Y. Hu and Kyunghyun Cho},
      year={2025},
      eprint={2507.00885},
      archivePrefix={arXiv},
      primaryClass={cs.CL},
      url={https://arxiv.org/abs/2507.00885}, 
}

@misc{chen2023skillitdatadrivenskillsframework,
      title={Skill-it! A Data-Driven Skills Framework for Understanding and Training Language Models}, 
      author={Mayee F. Chen and Nicholas Roberts and Kush Bhatia and Jue Wang and Ce Zhang and Frederic Sala and Christopher Ré},
      year={2023},
      eprint={2307.14430},
      archivePrefix={arXiv},
      primaryClass={cs.CL},
      url={https://arxiv.org/abs/2307.14430}, 
}

@misc{wei2022emergentabilitieslargelanguage,
      title={Emergent Abilities of Large Language Models}, 
      author={Jason Wei and Yi Tay and Rishi Bommasani and Colin Raffel and Barret Zoph and Sebastian Borgeaud and Dani Yogatama and Maarten Bosma and Denny Zhou and Donald Metzler and Ed H. Chi and Tatsunori Hashimoto and Oriol Vinyals and Percy Liang and Jeff Dean and William Fedus},
      year={2022},
      eprint={2206.07682},
      archivePrefix={arXiv},
      primaryClass={cs.CL},
      url={https://arxiv.org/abs/2206.07682}, 
}

@misc{snell2024predictingemergentcapabilitiesfinetuning,
      title={Predicting Emergent Capabilities by Finetuning}, 
      author={Charlie Snell and Eric Wallace and Dan Klein and Sergey Levine},
      year={2024},
      eprint={2411.16035},
      archivePrefix={arXiv},
      primaryClass={cs.LG},
      url={https://arxiv.org/abs/2411.16035}, 
}

@misc{olsson2022incontextlearninginductionheads,
      title={In-context Learning and Induction Heads}, 
      author={Catherine Olsson and Nelson Elhage and Neel Nanda and Nicholas Joseph and Nova DasSarma and Tom Henighan and Ben Mann and Amanda Askell and Yuntao Bai and Anna Chen and Tom Conerly and Dawn Drain and Deep Ganguli and Zac Hatfield-Dodds and Danny Hernandez and Scott Johnston and Andy Jones and Jackson Kernion and Liane Lovitt and Kamal Ndousse and Dario Amodei and Tom Brown and Jack Clark and Jared Kaplan and Sam McCandlish and Chris Olah},
      year={2022},
      eprint={2209.11895},
      archivePrefix={arXiv},
      primaryClass={cs.LG},
      url={https://arxiv.org/abs/2209.11895}, 
}

@misc{lee2025distinctcomputationsemergecompositional,
      title={Distinct Computations Emerge From Compositional Curricula in In-Context Learning}, 
      author={Jin Hwa Lee and Andrew K. Lampinen and Aaditya K. Singh and Andrew M. Saxe},
      year={2025},
      eprint={2506.13253},
      archivePrefix={arXiv},
      primaryClass={cs.LG},
      url={https://arxiv.org/abs/2506.13253}, 
}

@misc{zhang2026saddletosaddledynamicsexplainssimplicity,
      title={Saddle-to-Saddle Dynamics Explains A Simplicity Bias Across Neural Network Architectures}, 
      author={Yedi Zhang and Andrew Saxe and Peter E. Latham},
      year={2026},
      eprint={2512.20607},
      archivePrefix={arXiv},
      primaryClass={cs.LG},
      url={https://arxiv.org/abs/2512.20607}, 
}

@inproceedings{
yin2025which,
title={Which Attention Heads Matter for In-Context Learning?},
author={Kayo Yin and Jacob Steinhardt},
booktitle={Forty-second International Conference on Machine Learning},
year={2025},
url={https://openreview.net/forum?id=C7XmEByCFv}
}

@techreport{ekstrom1976kit,
  title     = {Manual for Kit of Factor-Referenced Cognitive Tests},
  author    = {Ekstrom, Ruth B. and French, John W. and Harman, Harry H. and Dermen, Diran},
  institution = {Educational Testing Service},
  address   = {Princeton, NJ},
  year      = {1976}
}

@inproceedings{wang2023interpretability,
  title     = {Interpretability in the Wild: A Circuit for Indirect Object Identification in {GPT}-2 Small},
  author    = {Wang, Kevin and Variengien, Alexandre and Conmy, Arthur and Shlegeris, Buck and Steinhardt, Jacob},
  booktitle = {International Conference on Learning Representations},
  year      = {2023},
  url       = {https://arxiv.org/abs/2211.00593}
}

@inproceedings{chen2024parallel,
  title     = {Parallel Structures in Pre-training Data Yield In-Context Learning},
  author    = {Chen, Yanda and Zhao, Chen and Yu, Zhou and McKeown, Kathleen and He, He},
  booktitle = {Proceedings of the 62nd Annual Meeting of the Association for Computational Linguistics (Volume 1: Long Papers)},
  pages     = {8582--8592},
  year      = {2024},
  address   = {Bangkok, Thailand},
  publisher = {Association for Computational Linguistics},
  url       = {https://aclanthology.org/2024.acl-long.465}
}

@inproceedings{feucht2025dualroute,
  title     = {The Dual-Route Model of Induction},
  author    = {Feucht, Sheridan and Todd, Eric and Wallace, Byron and Bau, David},
  booktitle = {Second Conference on Language Modeling},
  year      = {2025},
  url       = {https://arxiv.org/abs/2504.03022}
}

@inproceedings{liu-etal-2021-probing-across,
    title = "Probing Across Time: What Does {R}o{BERT}a Know and When?",
    author = "Liu, Leo Z.  and
      Wang, Yizhong  and
      Kasai, Jungo  and
      Hajishirzi, Hannaneh  and
      Smith, Noah A.",
    editor = "Moens, Marie-Francine  and
      Huang, Xuanjing  and
      Specia, Lucia  and
      Yih, Scott Wen-tau",
    booktitle = "Findings of the Association for Computational Linguistics: EMNLP 2021",
    month = nov,
    year = "2021",
    address = "Punta Cana, Dominican Republic",
    publisher = "Association for Computational Linguistics",
    url = "https://aclanthology.org/2021.findings-emnlp.71/",
    doi = "10.18653/v1/2021.findings-emnlp.71",
    pages = "820--842"
}
\bibliographystyle{colm2026_conference}

\onecolumn
\appendix

\section{LLM Usage Disclosure}

Claude Opus 4.6 and Sonnet 4.6 were used to format tables and conduct minor writing edits. All outputs were reviewed, and verified by the authors. 

\section{Extended Related Work}
\label{appendix:related_work_extended}


\paragraph{Skill Emergence and Scaling Laws.}
Theoretical work has sought to explain how capabilities emerge with scale.
\citet{arora2023theoryemergencecomplexskills} propose that scaling laws arise from slingshot generalization, where competence at $k$-tuples of skills emerges at the same rate as elementary skills themselves.
Similarly, \citet{michaud2023the} introduce the quanta hypothesis, modeling skills as discrete units whose power-law frequency distribution explains smooth scaling curves.
Both theories predict that complex behaviors emerge from simpler building blocks, but leave open the question of what these building blocks are and how they compose in practice.
Our work provides empirical grounding for these theories by tracking probe tasks designed to be compositionally combined, finding that compositional skills reliably emerge after their constituent components.

\paragraph{Skill Evaluation and Structure.}
Several approaches characterize LLM capabilities through evaluation-time analysis.
\citet{burnell2023revealingstructurelanguagemodel} apply factor analysis across 29 models and 27 tasks, finding three latent factors, reasoning, comprehension, and language modeling, that explain performance variation; \citet{maimon2025iqtestllmsevaluation} scale this psychometric approach to 60 models and 44 tasks, identifying eight core skills.
Beyond identifying skills, \citet{yu2024skillmix} directly test compositional ability by evaluating whether models can combine $k$-tuples of language skills in novel ways.
\citet{polo2025sloth} unify these perspectives through skill-based scaling laws where performance is driven by low-dimensional latent skills.
These works analyze fully trained models; we complement them by studying \emph{how skills develop during pretraining} and linking emergence order to representational structure.

\paragraph{Training Dynamics and Phase Transitions.}
Understanding what models learn during training has gained increasing attention.
\citet{chen2023sudden} identify sudden drops in loss corresponding to syntax acquisition and other phase transitions; \citet{kangaslahti2025hidden} show that such breakthroughs occur frequently but are obscured by aggregate loss metrics.
\citet{van2025polypythias} release 50 additional training runs of Pythia models, finding consistent learning phases across seeds and sizes.
Other work examines specific capabilities: \citet{sun-dredze-2025-amuro} investigate how downstream performance develops across pretraining checkpoints, \citet{ge2025evolution} track feature evolution using sparse dictionary learning, and \citet{mishranext} show that mathematical skills emerge in an order correlated with human curriculum despite random data ordering.
Our work contributes to this literature by demonstrating that emergence orderings are stable across model families and can be predicted from representational geometry.

\paragraph{Representations for Task Understanding.}
Mechanistic interpretability has revealed compact representations of tasks within model activations.
Both \citet{todd2024function} and \citet{hendel-etal-2023-context} discover that in-context learning compresses task demonstrations into single directions, termed function vectors and task vectors respectively, which can trigger task execution even in zero-shot settings.
Subsequent work explores the scope of this phenomenon: \citet{todd2026incontextalgebra} extend it to symbolic reasoning with variable-based tokens, while \citet{khandelwal2025languagemodelscomposefunctions} investigate compositional tasks and find both compositional and direct processing mechanisms.
We build on this line of work by using residual-stream representations to predict learning trajectories of compositional tasks, connecting representational geometry to training dynamics.
This complements recent work by \citet{prasad2026featuresrewardsscalablesupervision} that uses interpretable features for training monitoring in RL.

\section{Full list of Elemental and Composite Tasks}
\label{appendix:all_tasks}

We provide a full list of tasks, categorized into reasoning types, in \autoref{tab:all-elemental-tasks} and \autoref{tab:all-compositional-tasks}. TextFRCT tasks are taken from the psychometrics literature \citep{ekstrom1976kit}, while other tasks have been studied or were inspired by works in interpretability literature \citep{wang2023interpretability, todd2024function, chen2024parallel, feucht2025dualroute}.

\section{Full list of Elemental and Composite Tasks}
\label{appendix:all_tasks}

We provide a full list of tasks, categorized into reasoning types, in \autoref{tab:all-elemental-tasks} and \autoref{tab:all-compositional-tasks}. TextFRCT tasks are taken from the psychometrics literature \citep{ekstrom1976kit}, while other tasks have been studied or were inspired by works in interpretability literature \citep{wang2023interpretability, todd2024function, chen2024parallel, feucht2025dualroute}. 


{\scriptsize
\setlength{\tabcolsep}{3pt}
\renewcommand{\arraystretch}{0.96}
\begin{longtable}{p{4.25cm}p{0.75cm}p{5.85cm}p{1.15cm}}
\caption{All elemental tasks in the evaluation suite with representative examples.}
\label{tab:all-elemental-tasks} \\
\toprule
\textbf{Task} & \textbf{N} & \textbf{Input} & \textbf{Output} \\
\midrule
\endfirsthead
\multicolumn{4}{l}{\textit{Table~\ref{tab:all-elemental-tasks} continued from previous page}} \\[0.3em]
\toprule
\textbf{Task} & \textbf{N} & \textbf{Input} & \textbf{Output} \\
\midrule
\endhead
\midrule
\multicolumn{4}{r}{\textit{Continued on next page}} \\
\endfoot
\bottomrule
\endlastfoot
%
\multicolumn{4}{l}{\textit{String Operations}} \\[0.2em]
\texttt{copying} & 20 & gTpigTHK & gTpigTHK \\
\addlinespace[0.3em]
\texttt{token\_reversal} & 20 & cat & tac \\
\addlinespace[0.3em]
\texttt{string\_analogy} & 10 & abc $\to$ abd, ijk $\to$ ? & ijl \\
\addlinespace[0.3em]
\texttt{simple\_icl:uppercase} & 26 & b & B \\
\addlinespace[0.3em]
\texttt{simple\_icl:lowercase} & 26 & B & b \\
\addlinespace[0.3em]
\texttt{simple\_icl:first\_letter} & 190 & the cat went up the tree & t \\
\addlinespace[0.3em]
\texttt{simple\_icl:last\_letter} & 190 & the cat went up the tree & e \\
\addlinespace[0.8em]
%
\multicolumn{4}{l}{\textit{Morphology}} \\[0.2em]
\texttt{simple\_icl:present\_to\_gerund} & 179 & run & running \\
\addlinespace[0.3em]
\texttt{simple\_icl:singular\_to\_plural} & 165 & child & children \\
\addlinespace[0.8em]
%
\multicolumn{4}{l}{\textit{Translation}} \\[0.2em]
\texttt{simple\_icl:translate\_eng\_fr} & 173 & hello & bonjour \\
\addlinespace[0.3em]
\texttt{simple\_icl:translate\_fr\_eng} & 175 & bonjour & hello \\
\addlinespace[0.3em]
\texttt{simple\_icl:translate\_eng\_sp} & 178 & hello & hola \\
\addlinespace[0.3em]
\texttt{simple\_icl:translate\_sp\_eng} & 178 & hola & hello \\
\addlinespace[0.8em]
%
\multicolumn{4}{l}{\textit{World Knowledge}} \\[0.2em]
\texttt{simple\_icl:country\_to\_capital} & 184 & Afghanistan & Kabul \\
\addlinespace[0.3em]
\texttt{simple\_icl:country\_to\_currency} & 198 & United States & Dollar \\
\addlinespace[0.8em]
%
\multicolumn{4}{l}{\textit{Arithmetic}} \\[0.2em]
\texttt{basic\_arithmetic} & 10 & What is 5 + 3? & 8 \\
\addlinespace[0.3em]
\texttt{math} & 20 & 4 * 1 & 4 \\
\addlinespace[0.3em]
\texttt{multistep\_arithmetic:two\_step} & 20 & 3 + 4, then multiply by 2 & 14 \\
\addlinespace[0.3em]
\texttt{multistep\_arithmetic:three\_step} & 20 & Start with 10, subtract 3, then multiply by 4 & 28 \\
\addlinespace[0.3em]
\texttt{textfrct:RG1} & 30 & In general, brass is made of two parts copper to one part zinc. How many pounds of zinc are needed to produce 45 pounds of brass? \textit{(MCQ)} & B \\
\addlinespace[0.3em]
\texttt{textfrct:RG2} & 30 & Recipe A uses 1.5 cups of sugar; Recipe B uses 2. Making 8 cakes, how many fewer cups does Recipe A require? \textit{(MCQ)} & E \\
\addlinespace[0.3em]
\texttt{textfrct:RG3} & 30 & There are 4 quarts in a gallon and 4 cups in a quart. How many cups are in a gallon? \textit{(MCQ)} & C \\
\addlinespace[0.8em]
%
\multicolumn{4}{l}{\textit{Logic}} \\[0.2em]
\texttt{logical\_ops:negation} & 12 & Statement: All robots can move.\newline Candidate: Some robots cannot move.\newline Is this a correct logical negation? & True \\
\addlinespace[0.3em]
\texttt{logical\_ops:conjunction} & 12 & Fact A is True. Fact B is True.\newline Claim: A AND B. Is the claim true? & True \\
\addlinespace[0.3em]
\texttt{logical\_ops:conditional} & 12 & Rule: If it rains, the ground gets wet.\newline Fact: It rains. Does the conclusion follow? & True \\
\addlinespace[0.3em]
\texttt{textfrct:RL1} & 30 & All birds have purple tails. All cats are birds. Therefore all cats have purple tails. \textit{(MCQ: correct/incorrect)} & G \\
\addlinespace[0.3em]
\texttt{textfrct:RL3} & 20 & More fatal accidents occur on highways after dark than during daylight hours. \textit{(MCQ: which conclusion follows?)} & 3 \\
\addlinespace[0.3em]
\texttt{textfrct:RL4} & 24 & \textit{ICL ex.: black sheep = dag kip; white dog = tin bud; black cow = dag stam}\newline Query: white sheep = ? \textit{(MCQ)} & 2 \\
\addlinespace[0.8em]
%
\multicolumn{4}{l}{\textit{Reading Comprehension}} \\[0.2em]
\texttt{fact\_extraction:extract\_entity} & 20 & Passage: ``Alice gave five apples to Bob at the park.'' Who received the apples? & Bob \\
\addlinespace[0.3em]
\texttt{fact\_extraction:extract\_number} & 20 & Passage: ``John gave 5 apples to Mary on Tuesday.'' How many apples did John give? & 5 \\
\addlinespace[0.3em]
\texttt{fact\_extraction:extract\_location} & 20 & Passage: ``The cat sat on the red mat in the kitchen.'' Where is the mat? & the kitchen \\
\addlinespace[0.3em]
\texttt{coreference:pronoun\_simple} & 20 & ``Alice told Bob that she would be late.'' Who does ``she'' refer to? & Alice \\
\addlinespace[0.3em]
\texttt{coreference:pronoun\_hard} & 20 & ``The trophy didn't fit in the suitcase because it was too big.'' What was too big? & the trophy \\
\addlinespace[0.3em]
\texttt{ignoring\_context} & 5 & Some text here. X = 5. More text.\newline Question: What is X? & 5 \\
\addlinespace[0.3em]
\texttt{ioi\_task} & 1000 & \textit{Instr.: Identify who should be referenced.}\newline Then, Henry and Phil had a lot of fun at the harbor. Henry gave a basket to & \texttt{[Phil, Henry]} \\
\addlinespace[0.3em]
\texttt{part\_of\_speech} & 15 & The cat is in the house. The part of speech for ``cat'' is \_ & noun \\
\addlinespace[0.8em]
%
\multicolumn{4}{l}{\textit{Verbal Closure (FRCT)}} \\[0.2em]
\texttt{textfrct:CV1} & 50 & erte & tree, rete \\
\addlinespace[0.3em]
\texttt{textfrct:CV2} & 40 & EZIRTMODSLOWTSEXQILNECKBWOCJAKX & SLOW, NECK \\
\addlinespace[0.3em]
\texttt{textfrct:CV3} & 36 & \_tam\_ & stamp \\
\addlinespace[0.8em]
%
\multicolumn{4}{l}{\textit{Induction (FRCT)}} \\[0.2em]
\texttt{textfrct:I1} & 30 & \textit{Instr.: One of the five letter sets does NOT follow the same pattern as the others. Find it.}\newline 1.~QPPQ \quad 2.~HGHH \quad 3.~TTTU \quad 4.~DDDE \quad 5.~MLMM & 1 \\[1.6em]
\addlinespace[0.3em]
\texttt{textfrct:I2} & 28 & \textit{Instr.: Each row marks one position with `x'. Identify the pattern and find the correct position in row 5.}\newline
\texttt{------- x------- ---- --}\newline
\texttt{---- -x--- -- --- ------}\newline
\texttt{--------------- --x-----}\newline
\texttt{-------- ---x-----------}\newline
\texttt{----1 2---3-- 4---5-----} & 3 \\[4.0em]
\addlinespace[0.8em]
%
\multicolumn{4}{l}{\textit{Associative Memory (FRCT)}} \\[0.2em]
\texttt{textfrct:MA2} & 30 & \textit{Instr.: Memorize 30 word--number pairs, then answer retrieval queries.}\newline Query: What number corresponds to `coat'? & 49 \\
\addlinespace[0.3em]
\texttt{textfrct:MA3} & 30 & \textit{Instr.: Memorize 30 first--last name pairs, then answer retrieval queries.}\newline Query: Last name: Nichols & Edward \\
\addlinespace[0.8em]
%
\multicolumn{4}{l}{\textit{Verbal Comprehension (FRCT)}} \\[0.2em]
\texttt{textfrct:V1} & 36 & \textit{Instr.: Choose the best definition (MCQ).}\newline `airtight': (1)~firm \; (2)~light \; (3)~hermetically sealed \; (4)~plane sick & 3 \\
\addlinespace[0.3em]
\texttt{textfrct:V2} & 36 & \textit{Instr.: Choose the best definition (MCQ).}\newline `handicraft': (1)~cunning \; (2)~fast boat \; (3)~utility \; (4)~manual skill \; (5)~guild & 4 \\
\addlinespace[0.3em]
\texttt{textfrct:V3} & 48 & \textit{Instr.: Choose the best definition (MCQ).}\newline `cottontail': (1)~squirrel \; (2)~poplar \; (3)~boa \; (4)~marshy plant \; (5)~rabbit & 5 \\
\addlinespace[0.3em]
\texttt{textfrct:V4} & 36 & \textit{Instr.: Choose the best definition (MCQ).}\newline `mumble': (1)~speak indistinctly \; (2)~complain \; (3)~handle awkwardly \; (4)~fall \; (5)~tear apart & 1 \\
\addlinespace[0.3em]
\texttt{textfrct:V5} & 36 & \textit{Instr.: Choose the best definition (MCQ).}\newline `rancor': (1)~forbearance \; (2)~ridicule \; (3)~malice \; (4)~bravery & 3 \\
\end{longtable}
}


{\scriptsize
\setlength{\tabcolsep}{3pt}
\renewcommand{\arraystretch}{0.96}
\begin{longtable}{p{5.25cm}p{0.75cm}p{4.45cm}p{1.15cm}}
\caption{All compositional tasks in the evaluation suite with representative examples.}
\label{tab:all-compositional-tasks} \\
\toprule
\textbf{Task} & \textbf{N} & \textbf{Input} & \textbf{Output} \\
\midrule
\endfirsthead
\multicolumn{4}{l}{\textit{Table~\ref{tab:all-compositional-tasks} continued from previous page}} \\[0.3em]
\toprule
\textbf{Task} & \textbf{N} & \textbf{Input} & \textbf{Output} \\
\midrule
\endhead
\midrule
\multicolumn{4}{r}{\textit{Continued on next page}} \\
\endfoot
\bottomrule
\endlastfoot
%
\multicolumn{4}{l}{\textit{Morphology $\times$ String Operation}} \\[0.2em]
\texttt{compositional:gerund\_lower} & 178 & RUN & running \\
\addlinespace[0.3em]
\texttt{compositional:gerund\_upper} & 178 & run & RUNNING \\
\addlinespace[0.3em]
\texttt{compositional:gerund\_reverse} & 178 & run & gninnur \\
\addlinespace[0.3em]
\texttt{compositional:gerund\_upper\_reverse} & 178 & run & GNINNUR \\
\addlinespace[0.3em]
\texttt{compositional:plural\_lower} & 165 & CHILD & children \\
\addlinespace[0.3em]
\texttt{compositional:plural\_upper} & 165 & child & CHILDREN \\
\addlinespace[0.3em]
\texttt{compositional:plural\_reverse} & 165 & child & nerdlihc \\
\addlinespace[0.3em]
\texttt{compositional:plural\_upper\_reverse} & 165 & child & NERDLIHC \\
\addlinespace[0.8em]
%
\multicolumn{4}{l}{\textit{Translation $\times$ String Operation}} \\[0.2em]
\texttt{compositional:translate\_eng\_fr\_first} & 173 & hello & b \\
\addlinespace[0.3em]
\texttt{compositional:translate\_eng\_fr\_last} & 173 & hello & r \\
\addlinespace[0.3em]
\texttt{compositional:translate\_eng\_fr\_lower} & 173 & HELLO & bonjour \\
\addlinespace[0.3em]
\texttt{compositional:translate\_eng\_fr\_reverse} & 173 & hello & ruojnob \\
\addlinespace[0.3em]
\texttt{compositional:translate\_eng\_fr\_upper} & 173 & hello & BONJOUR \\
\addlinespace[0.3em]
\texttt{compositional:translate\_eng\_fr\_upper\_reverse} & 173 & hello & RUOJNOB \\
\addlinespace[0.3em]
\texttt{compositional:translate\_eng\_sp\_first} & 178 & hello & h \\
\addlinespace[0.3em]
\texttt{compositional:translate\_eng\_sp\_last} & 178 & hello & a \\
\addlinespace[0.3em]
\texttt{compositional:translate\_eng\_sp\_lower} & 178 & HELLO & hola \\
\addlinespace[0.3em]
\texttt{compositional:translate\_eng\_sp\_reverse} & 178 & hello & aloh \\
\addlinespace[0.3em]
\texttt{compositional:translate\_eng\_sp\_upper} & 178 & hello & HOLA \\
\addlinespace[0.3em]
\texttt{compositional:translate\_eng\_sp\_upper\_reverse} & 178 & hello & ALOH \\
\addlinespace[0.3em]
\texttt{compositional:translate\_fr\_eng\_first} & 171 & bonjour & h \\
\addlinespace[0.3em]
\texttt{compositional:translate\_fr\_eng\_last} & 171 & bonjour & o \\
\addlinespace[0.3em]
\texttt{compositional:translate\_fr\_eng\_lower} & 171 & BONJOUR & hello \\
\addlinespace[0.3em]
\texttt{compositional:translate\_fr\_eng\_reverse} & 171 & bonjour & olleh \\
\addlinespace[0.3em]
\texttt{compositional:translate\_fr\_eng\_upper} & 171 & bonjour & HELLO \\
\addlinespace[0.3em]
\texttt{compositional:translate\_sp\_eng\_first} & 178 & hola & h \\
\addlinespace[0.3em]
\texttt{compositional:translate\_sp\_eng\_last} & 178 & hola & o \\
\addlinespace[0.3em]
\texttt{compositional:translate\_sp\_eng\_lower} & 178 & HOLA & hello \\
\addlinespace[0.3em]
\texttt{compositional:translate\_sp\_eng\_reverse} & 178 & hola & olleh \\
\addlinespace[0.3em]
\texttt{compositional:translate\_sp\_eng\_upper} & 178 & hola & HELLO \\
\addlinespace[0.8em]
%
\multicolumn{4}{l}{\textit{Case/Reversal Chains}} \\[0.2em]
\texttt{compositional:lower\_first} & 971 & AFGHANISTAN & a \\
\addlinespace[0.3em]
\texttt{compositional:lower\_last} & 971 & AFGHANISTAN & n \\
\addlinespace[0.3em]
\texttt{compositional:lower\_reverse} & 971 & AFGHANISTAN & natsinahgfa \\
\addlinespace[0.3em]
\texttt{compositional:upper\_first} & 971 & afghanistan & A \\
\addlinespace[0.3em]
\texttt{compositional:upper\_last} & 971 & afghanistan & N \\
\addlinespace[0.3em]
\texttt{compositional:upper\_reverse} & 971 & afghanistan & NATSINAHGFA \\
\addlinespace[0.3em]
\texttt{compositional:reverse\_first} & 971 & Afghanistan & n \\
\addlinespace[0.3em]
\texttt{compositional:reverse\_last} & 971 & Afghanistan & A \\
\end{longtable}
}

\newpage

\newpage

\section{Full learning trajectories by category}
\label{appendix:all_learning_traj}

\begin{figure}
    \centering
    \includegraphics[width=0.95\linewidth]{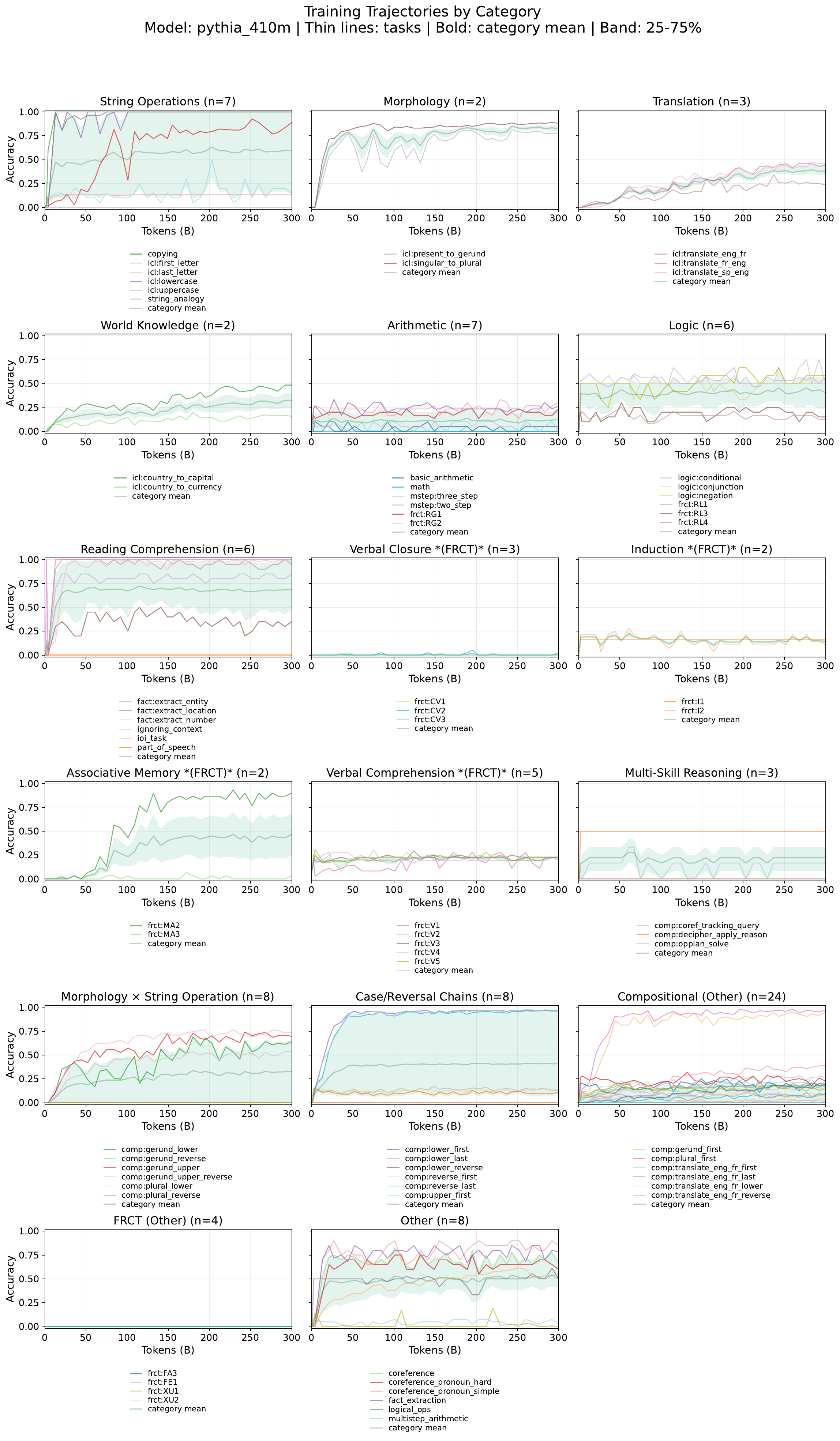}
    \caption{Complete trajectories for Pythia-410M over 300B tokens.}
    \label{fig:all_traj_pythia_410m}
\end{figure}

\begin{figure}
    \centering
    \includegraphics[width=0.95\linewidth]{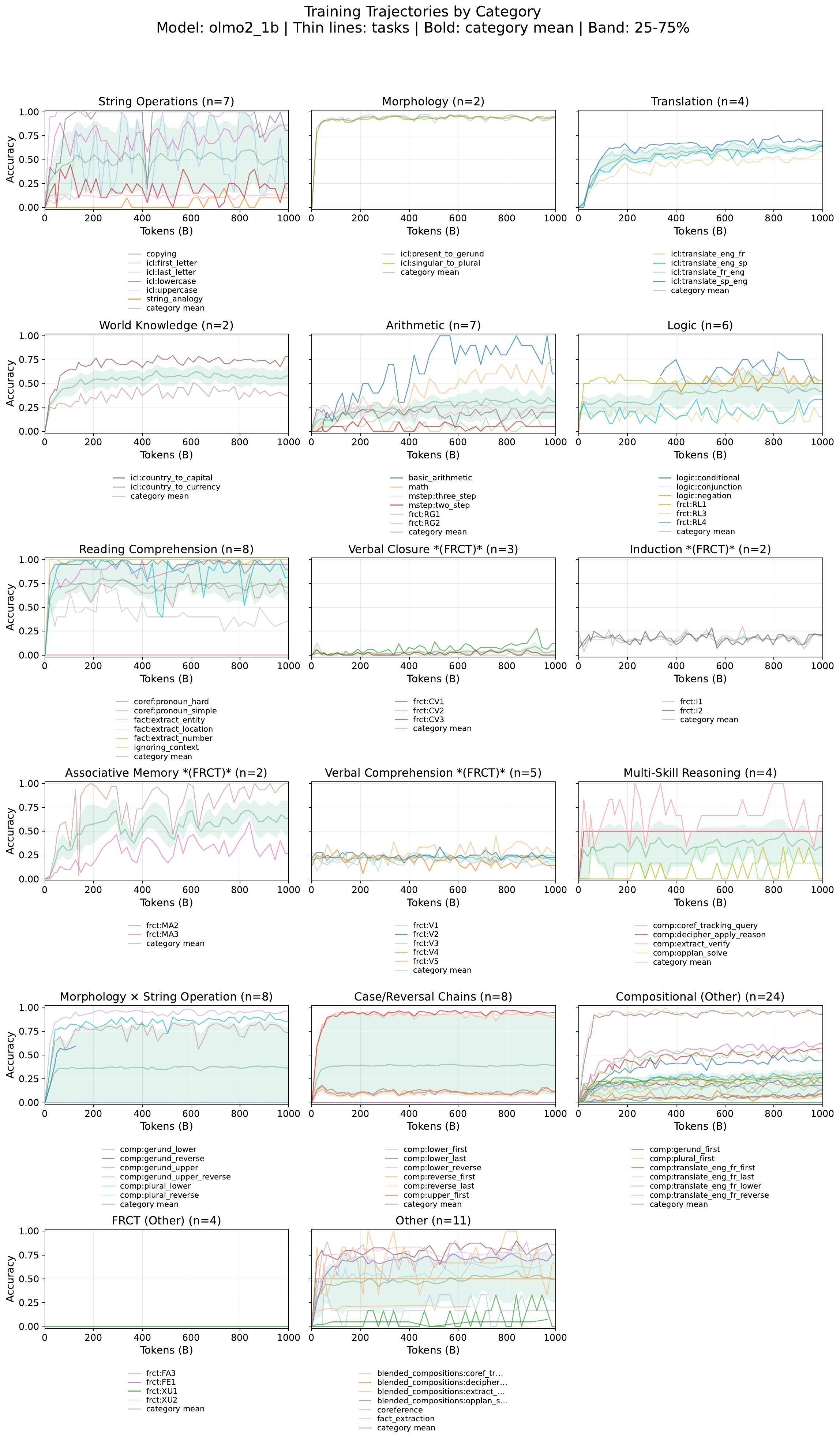}
    \caption{Complete trajectories for OLMo2-1B over 1T tokens.}
    \label{fig:all_traj_olmo2_1b}
\end{figure}

\begin{figure}
    \centering
    \includegraphics[width=0.95\linewidth]{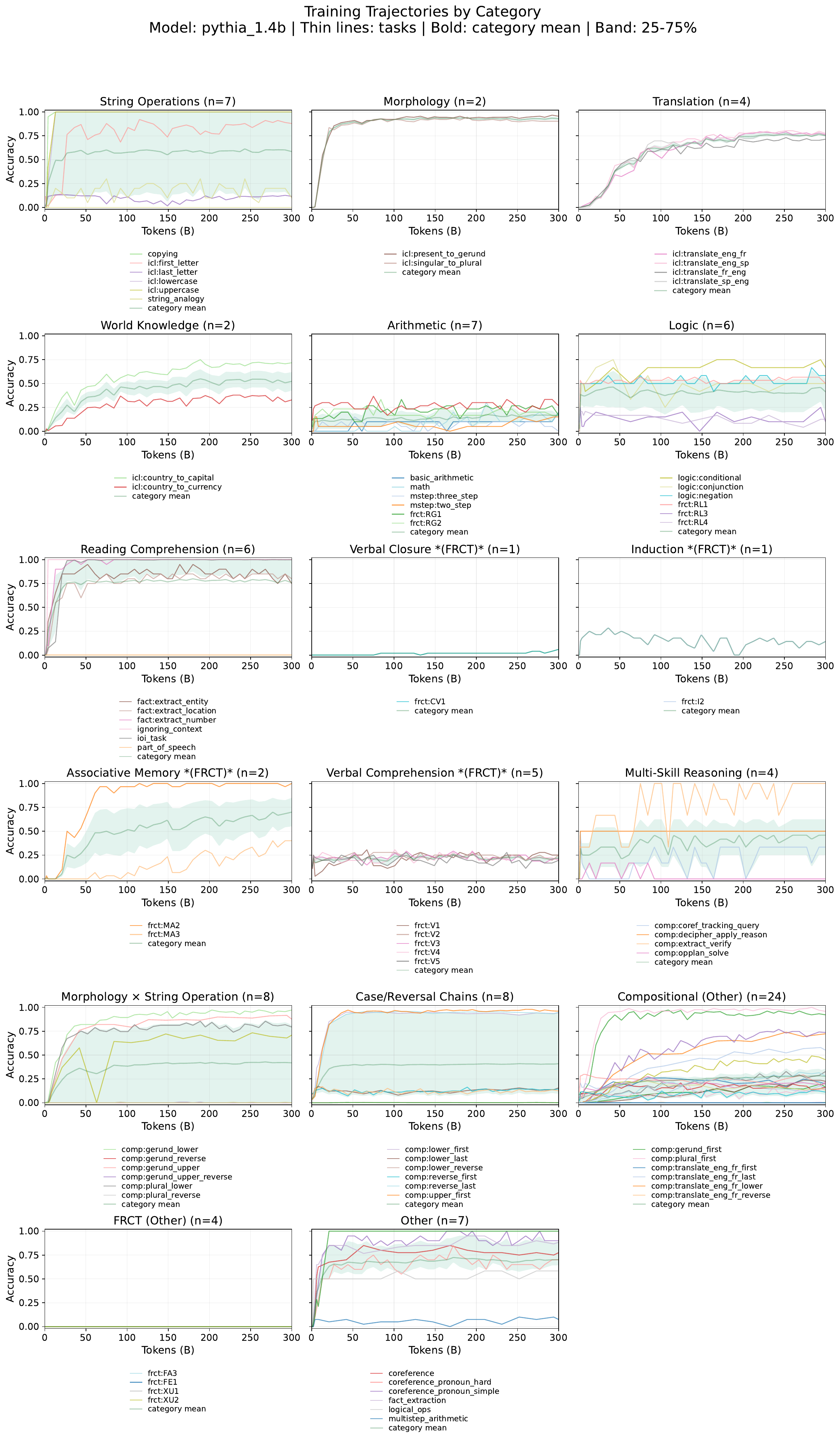}
    \caption{Complete trajectories for Pythia-1.4B over 300B tokens.}
    \label{fig:all_traj_pythia_1_4b}
\end{figure}

\begin{figure}
    \centering
    \includegraphics[width=0.95\linewidth]{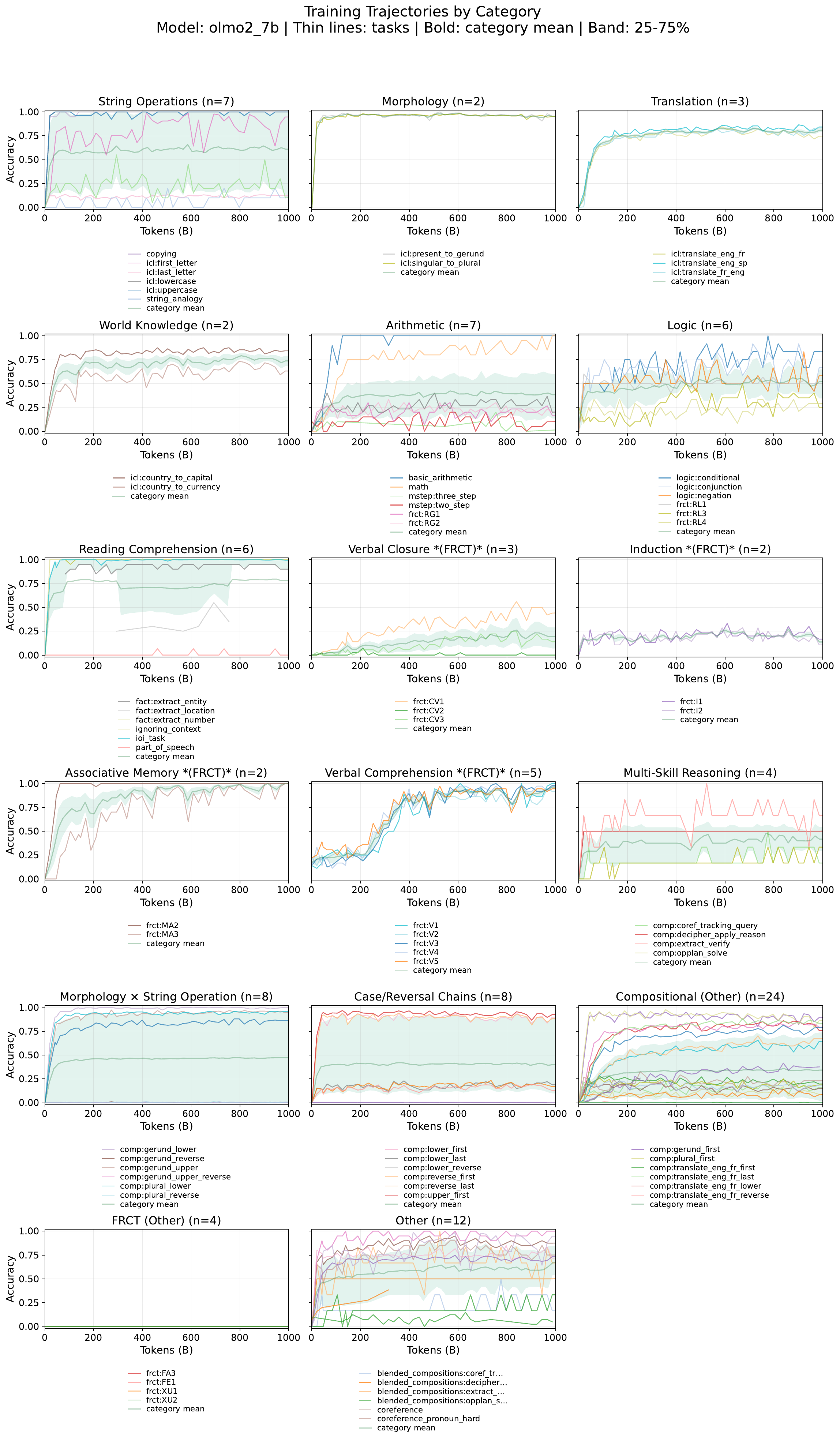}
    \caption{Complete trajectories for OLMo-2 7B over 1T tokens.}
    \label{fig:all_traj_olmo2_7b}
\end{figure}

\begin{figure}
    \centering
    \includegraphics[width=0.95\linewidth]{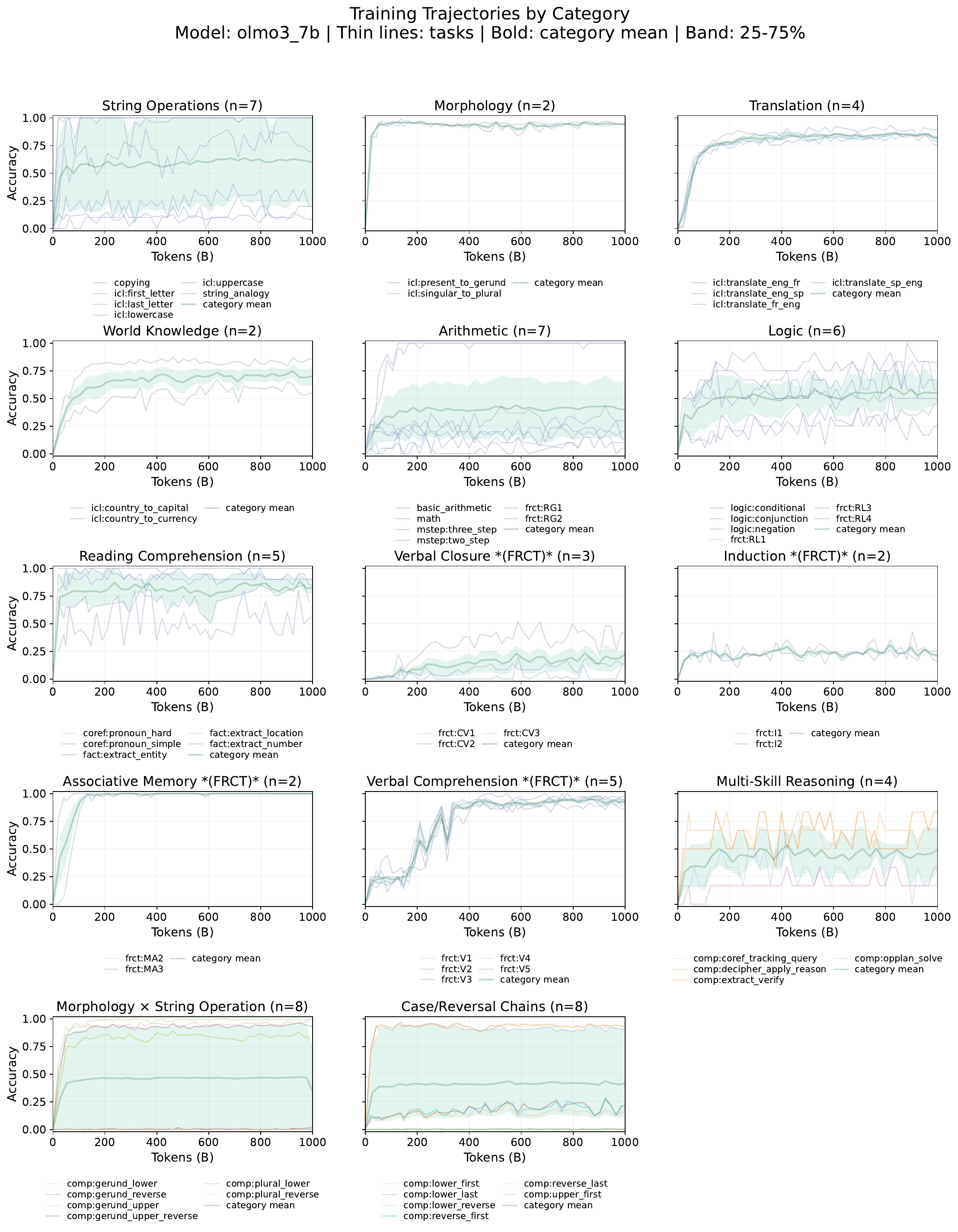}
    \caption{Complete trajectories for OLMo-3 7B over 1T tokens.}
    \label{fig:all_traj_olmo3_7b}
\end{figure}

\begin{figure}
    \centering
    \includegraphics[width=0.95\linewidth]{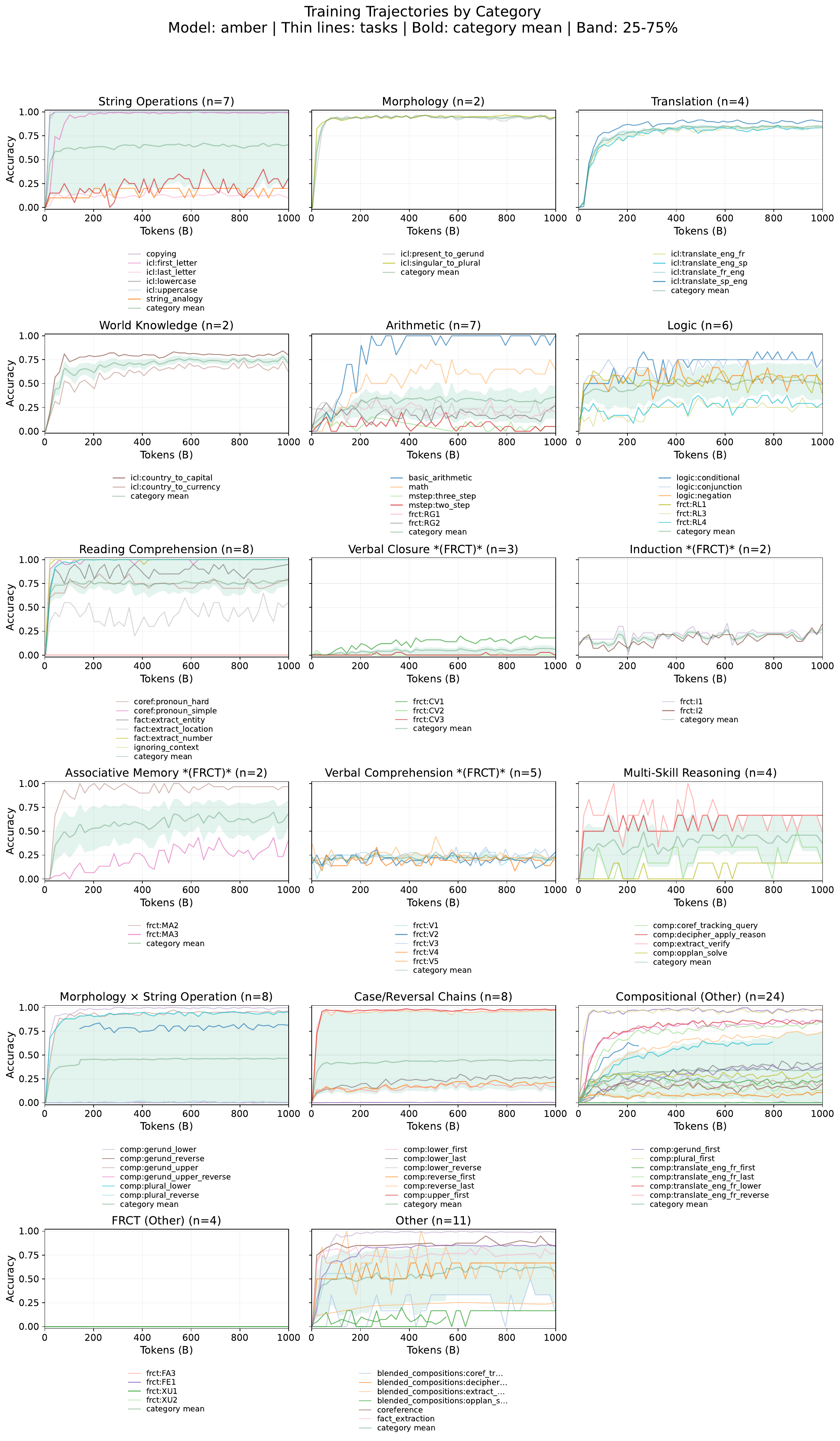}
    \caption{Complete trajectories for Amber (7B) over 1T tokens.}
    \label{fig:all_traj_amber_7b}
\end{figure}

\begin{figure}
    \centering
    \includegraphics[width=0.95\linewidth]{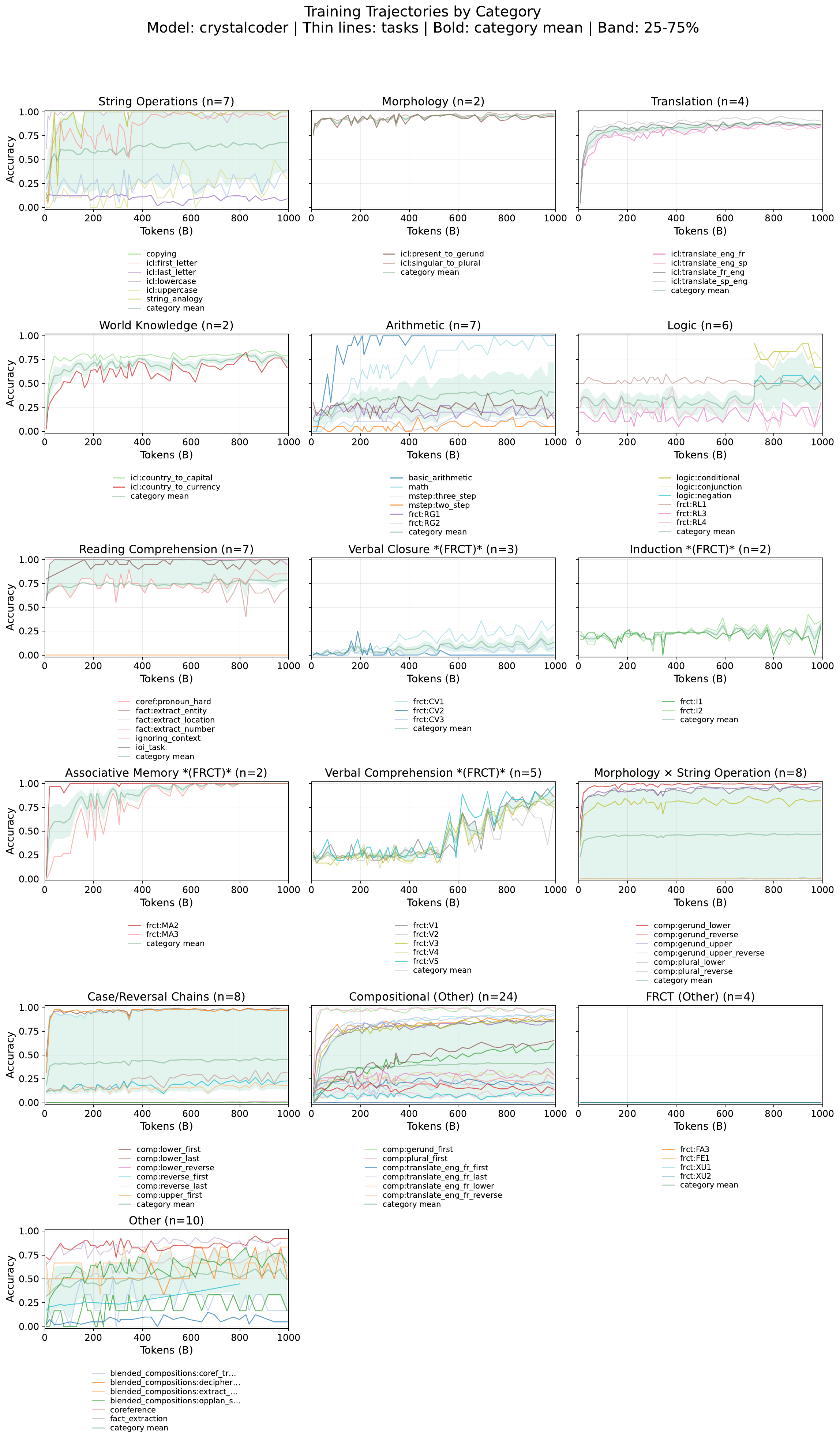}
    \caption{Complete trajectories for CrystalCoder (7B) over 1T tokens.}
    \label{fig:all_traj_crystal_7b}
\end{figure}

\begin{figure}
    \centering
    \includegraphics[width=0.95\linewidth]{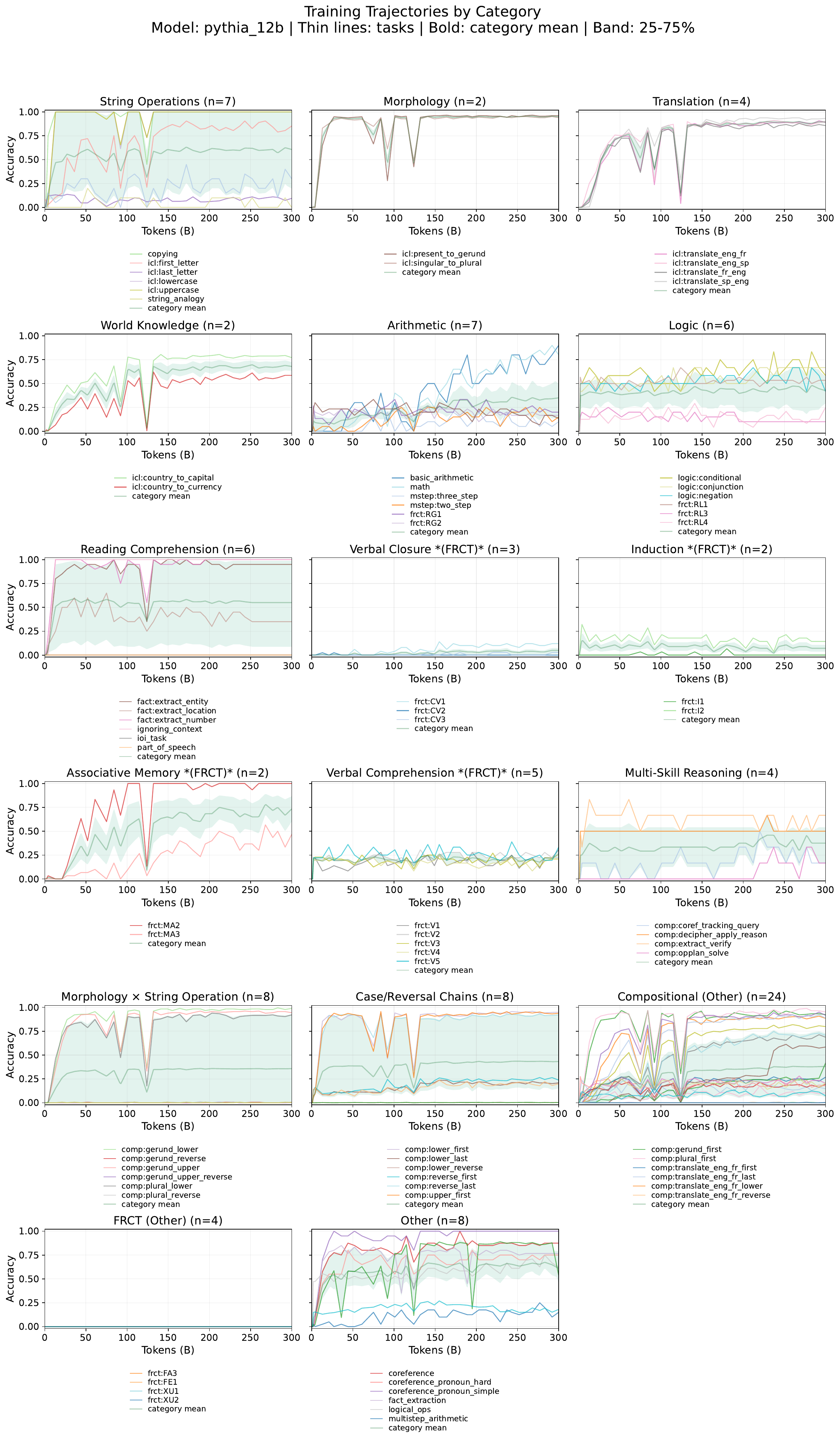}
    \caption{Complete trajectories for Pythia-12B over 300B tokens. Note that this model exhibits some instabilities compared to others.}
    \label{fig:all_traj_pythia_12b}
\end{figure}

\begin{figure}
    \centering
    \includegraphics[width=0.95\linewidth]{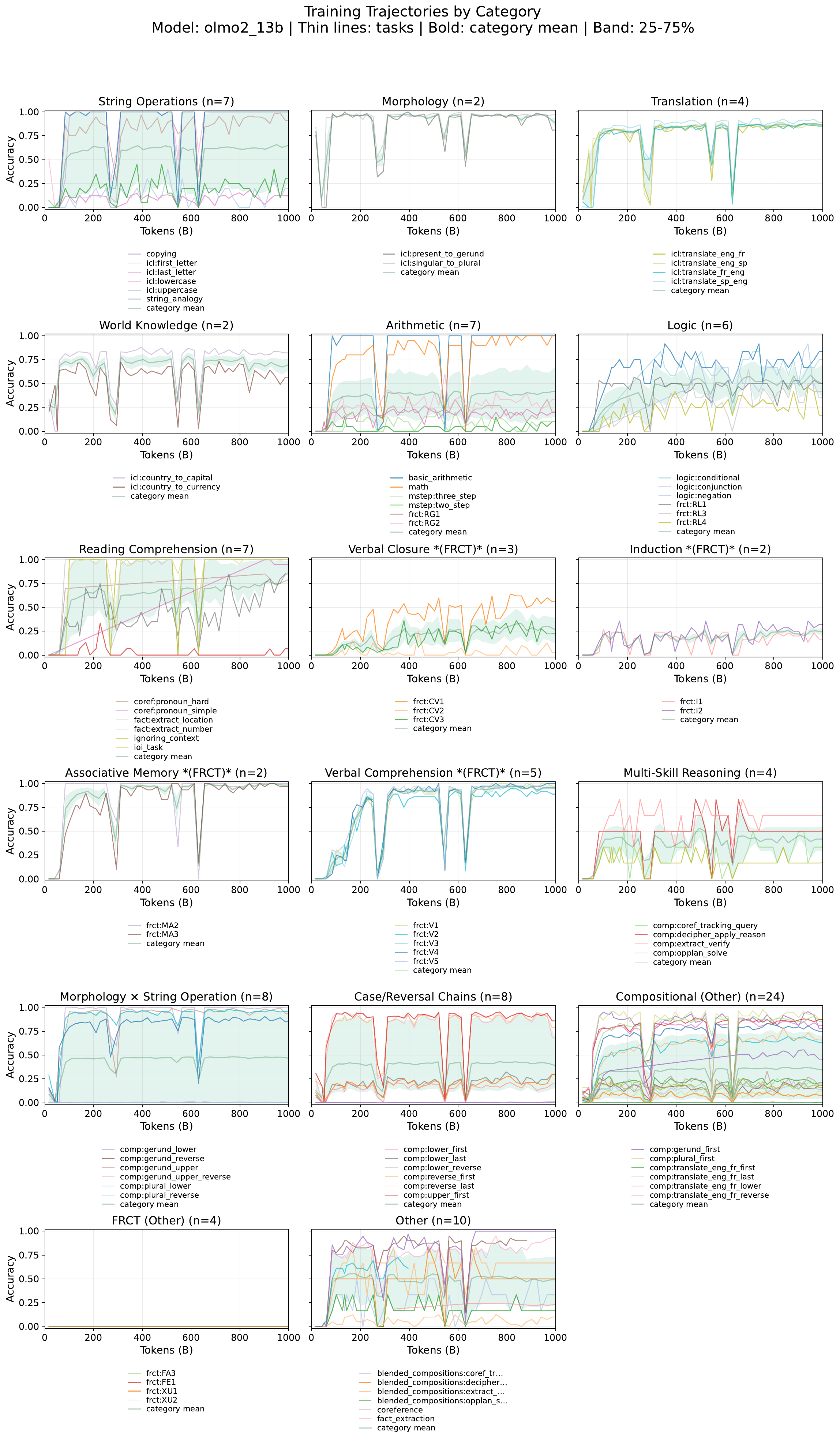}
    \caption{Complete trajectories for OLMo2-13b over 1T tokens. Note that this model exhibits some instabilities compared to others.}
    \label{fig:all_traj_olmo2_13b}
\end{figure}

Figures \ref{fig:all_traj_pythia_410m} -- \ref{fig:all_traj_olmo2_13b} show full learning trajectories of tasks for each model.

\section{Emergence Order Agreement Under Alternate Definitions}
\label{appendix:emergence_orders_alternate}

\begin{table}[h]
\centering
\caption{Summary of emergence ordering consistency under different definitions. Absolute thresholds yield substantially higher cross-model correlations than relative thresholds.}
\label{tab:emergence_defs_summary}
\small
\begin{tabular}{lcccc}
\toprule
\textbf{Definition} & \textbf{$n$ pairs} & \textbf{Mean $\rho$} & \textbf{Min $\rho$} & \textbf{Max $\rho$} \\
\midrule
Absolute threshold ($\theta = 0.5$) & 36 & 0.860 & 0.597 & 0.955 \\
Absolute threshold ($\theta = 0.8$, stable for 3 consecutive checkpoints) & 36 & 0.790 & 0.599 & 0.961 \\
\midrule
Relative threshold ($\alpha = 0.5$, fraction of max performance) & 36 & 0.528 & 0.085 & 0.866 \\
Relative threshold ($\alpha = 0.8$, fraction of max performance) & 36 & 0.500 & 0.077 & 0.773 \\
\bottomrule
\end{tabular}
\end{table}

 
\begin{table}[h]
\centering
\caption{Pairwise Spearman rank correlations: Absolute threshold ($\theta = 0.5$). Mean $\rho = 0.860$.}
\label{tab:spearman_abs05}
\small
\begin{tabular}{llccc}
\toprule
\textbf{Model A} & \textbf{Model B} & \textbf{$n$ tasks} & \textbf{$\rho$} & \textbf{$p$} \\
\midrule
\multicolumn{5}{l}{\textit{Within-family (OLMo-2)}} \\
OLMo2-1B & OLMo2-7B & 106 & 0.889 & $4.5 \times 10^{-37}$ \\
OLMo2-1B & OLMo2-13B & 105 & 0.832 & $4.5 \times 10^{-28}$ \\
OLMo2-7B & OLMo2-13B & 104 & 0.865 & $2.9 \times 10^{-32}$ \\
\midrule
\multicolumn{5}{l}{\textit{Within-family (Pythia)}} \\
Pythia-1.4B & Pythia-410M & 96 & 0.910 & $1.3 \times 10^{-37}$ \\
Pythia-1.4B & Pythia-12B & 98 & 0.909 & $3.7 \times 10^{-38}$ \\
Pythia-410M & Pythia-12B & 100 & 0.815 & $6.5 \times 10^{-25}$ \\
\midrule
\multicolumn{5}{l}{\textit{Within-family (LLM360)}} \\
Amber & Crystal & 102 & 0.905 & $6.8 \times 10^{-39}$ \\
\midrule
\multicolumn{5}{l}{\textit{Cross-family (OLMo-2 $\leftrightarrow$ OLMo-3)}} \\
OLMo2-1B & OLMo3-7B & 106 & 0.925 & $1.4 \times 10^{-45}$ \\
OLMo2-7B & OLMo3-7B & 105 & 0.918 & $4.7 \times 10^{-43}$ \\
OLMo2-13B & OLMo3-7B & 104 & 0.897 & $6.5 \times 10^{-38}$ \\
\midrule
\multicolumn{5}{l}{\textit{Cross-family (OLMo-2 $\leftrightarrow$ LLM360)}} \\
Amber & OLMo2-1B & 107 & 0.932 & $5.5 \times 10^{-48}$ \\
Amber & OLMo2-7B & 106 & 0.907 & $7.2 \times 10^{-41}$ \\
Amber & OLMo2-13B & 105 & 0.839 & $5.6 \times 10^{-29}$ \\
Crystal & OLMo2-1B & 102 & 0.913 & $1.3 \times 10^{-40}$ \\
Crystal & OLMo2-7B & 101 & 0.910 & $1.4 \times 10^{-39}$ \\
Crystal & OLMo2-13B & 100 & 0.889 & $6.3 \times 10^{-35}$ \\
\midrule
\multicolumn{5}{l}{\textit{Cross-family (OLMo-3 $\leftrightarrow$ LLM360)}} \\
Amber & OLMo3-7B & 106 & 0.918 & $1.6 \times 10^{-43}$ \\
Crystal & OLMo3-7B & 102 & 0.955 & $8.1 \times 10^{-55}$ \\
\midrule
\multicolumn{5}{l}{\textit{Cross-family (OLMo-2 $\leftrightarrow$ Pythia)}} \\
Pythia-1.4B & OLMo2-1B & 98 & 0.907 & $9.6 \times 10^{-38}$ \\
Pythia-1.4B & OLMo2-7B & 97 & 0.830 & $8.7 \times 10^{-26}$ \\
Pythia-1.4B & OLMo2-13B & 97 & 0.716 & $1.6 \times 10^{-16}$ \\
Pythia-410M & OLMo2-1B & 100 & 0.834 & $4.8 \times 10^{-27}$ \\
Pythia-410M & OLMo2-7B & 99 & 0.793 & $1.2 \times 10^{-22}$ \\
Pythia-410M & OLMo2-13B & 98 & 0.597 & $8.4 \times 10^{-11}$ \\
Pythia-12B & OLMo2-1B & 102 & 0.856 & $2.1 \times 10^{-30}$ \\
Pythia-12B & OLMo2-7B & 101 & 0.832 & $5.1 \times 10^{-27}$ \\
Pythia-12B & OLMo2-13B & 100 & 0.786 & $3.4 \times 10^{-22}$ \\
\midrule
\multicolumn{5}{l}{\textit{Cross-family (OLMo-3 $\leftrightarrow$ Pythia)}} \\
Pythia-1.4B & OLMo3-7B & 97 & 0.864 & $4.6 \times 10^{-30}$ \\
Pythia-410M & OLMo3-7B & 99 & 0.799 & $3.9 \times 10^{-23}$ \\
Pythia-12B & OLMo3-7B & 101 & 0.867 & $1.2 \times 10^{-31}$ \\
\midrule
\multicolumn{5}{l}{\textit{Cross-family (LLM360 $\leftrightarrow$ Pythia)}} \\
Amber & Pythia-1.4B & 98 & 0.935 & $7.4 \times 10^{-45}$ \\
Amber & Pythia-410M & 100 & 0.853 & $1.9 \times 10^{-29}$ \\
Amber & Pythia-12B & 102 & 0.930 & $3.6 \times 10^{-45}$ \\
Crystal & Pythia-1.4B & 93 & 0.824 & $3.8 \times 10^{-24}$ \\
Crystal & Pythia-410M & 96 & 0.751 & $1.2 \times 10^{-18}$ \\
Crystal & Pythia-12B & 97 & 0.850 & $3.3 \times 10^{-28}$ \\
\bottomrule
\end{tabular}
\end{table}

\begin{table}[h]
\centering
\caption{Pairwise Spearman rank correlations: Absolute threshold ($\theta = 0.8$, stable for 3 consecutive checkpoints). Mean $\rho = 0.790$.}
\label{tab:spearman_abs08stable}
\small
\begin{tabular}{llccc}
\toprule
\textbf{Model A} & \textbf{Model B} & \textbf{$n$ tasks} & \textbf{$\rho$} & \textbf{$p$} \\
\midrule
\multicolumn{5}{l}{\textit{Within-family (OLMo-2)}} \\
OLMo2-1B & OLMo2-7B & 106 & 0.718 & $4.5 \times 10^{-18}$ \\
OLMo2-1B & OLMo2-13B & 105 & 0.721 & $4.0 \times 10^{-18}$ \\
OLMo2-7B & OLMo2-13B & 104 & 0.934 & $2.0 \times 10^{-47}$ \\
\midrule
\multicolumn{5}{l}{\textit{Within-family (Pythia)}} \\
Pythia-1.4B & Pythia-410M & 96 & 0.824 & $6.1 \times 10^{-25}$ \\
Pythia-1.4B & Pythia-12B & 98 & 0.792 & $2.9 \times 10^{-22}$ \\
Pythia-410M & Pythia-12B & 100 & 0.689 & $2.3 \times 10^{-15}$ \\
\midrule
\multicolumn{5}{l}{\textit{Within-family (LLM360)}} \\
Amber & Crystal & 102 & 0.823 & $2.7 \times 10^{-26}$ \\
\midrule
\multicolumn{5}{l}{\textit{Cross-family (OLMo-2 $\leftrightarrow$ OLMo-3)}} \\
OLMo2-1B & OLMo3-7B & 106 & 0.743 & $7.9 \times 10^{-20}$ \\
OLMo2-7B & OLMo3-7B & 105 & 0.961 & $3.5 \times 10^{-59}$ \\
OLMo2-13B & OLMo3-7B & 104 & 0.953 & $8.3 \times 10^{-55}$ \\
\midrule
\multicolumn{5}{l}{\textit{Cross-family (OLMo-2 $\leftrightarrow$ LLM360)}} \\
Amber & OLMo2-1B & 107 & 0.785 & $1.6 \times 10^{-23}$ \\
Amber & OLMo2-7B & 106 & 0.877 & $6.2 \times 10^{-35}$ \\
Amber & OLMo2-13B & 105 & 0.875 & $3.0 \times 10^{-34}$ \\
Crystal & OLMo2-1B & 102 & 0.695 & $5.0 \times 10^{-16}$ \\
Crystal & OLMo2-7B & 101 & 0.838 & $8.8 \times 10^{-28}$ \\
Crystal & OLMo2-13B & 100 & 0.868 & $1.5 \times 10^{-31}$ \\
\midrule
\multicolumn{5}{l}{\textit{Cross-family (OLMo-3 $\leftrightarrow$ LLM360)}} \\
Amber & OLMo3-7B & 106 & 0.877 & $6.5 \times 10^{-35}$ \\
Crystal & OLMo3-7B & 102 & 0.853 & $6.1 \times 10^{-30}$ \\
\midrule
\multicolumn{5}{l}{\textit{Cross-family (OLMo-2 $\leftrightarrow$ Pythia)}} \\
Pythia-1.4B & OLMo2-1B & 98 & 0.883 & $2.3 \times 10^{-33}$ \\
Pythia-1.4B & OLMo2-7B & 97 & 0.693 & $3.6 \times 10^{-15}$ \\
Pythia-1.4B & OLMo2-13B & 97 & 0.669 & $7.3 \times 10^{-14}$ \\
Pythia-410M & OLMo2-1B & 100 & 0.779 & $1.4 \times 10^{-21}$ \\
Pythia-410M & OLMo2-7B & 99 & 0.686 & $4.4 \times 10^{-15}$ \\
Pythia-410M & OLMo2-13B & 98 & 0.629 & $4.2 \times 10^{-12}$ \\
Pythia-12B & OLMo2-1B & 102 & 0.777 & $7.4 \times 10^{-22}$ \\
Pythia-12B & OLMo2-7B & 101 & 0.786 & $2.1 \times 10^{-22}$ \\
Pythia-12B & OLMo2-13B & 100 & 0.843 & $4.4 \times 10^{-28}$ \\
\midrule
\multicolumn{5}{l}{\textit{Cross-family (OLMo-3 $\leftrightarrow$ Pythia)}} \\
Pythia-1.4B & OLMo3-7B & 97 & 0.752 & $7.3 \times 10^{-19}$ \\
Pythia-410M & OLMo3-7B & 99 & 0.689 & $3.1 \times 10^{-15}$ \\
Pythia-12B & OLMo3-7B & 101 & 0.839 & $5.7 \times 10^{-28}$ \\
\midrule
\multicolumn{5}{l}{\textit{Cross-family (LLM360 $\leftrightarrow$ Pythia)}} \\
Amber & Pythia-1.4B & 98 & 0.805 & $1.7 \times 10^{-23}$ \\
Amber & Pythia-410M & 100 & 0.758 & $7.4 \times 10^{-20}$ \\
Amber & Pythia-12B & 102 & 0.889 & $1.3 \times 10^{-35}$ \\
Crystal & Pythia-1.4B & 93 & 0.703 & $4.2 \times 10^{-15}$ \\
Crystal & Pythia-410M & 96 & 0.599 & $1.1 \times 10^{-10}$ \\
Crystal & Pythia-12B & 97 & 0.830 & $7.1 \times 10^{-26}$ \\
\bottomrule
\end{tabular}
\end{table}

\begin{table}[h]
\centering
\caption{Pairwise Spearman rank correlations: Relative threshold ($\alpha = 0.5$, fraction of max performance). Mean $\rho = 0.528$.}
\label{tab:spearman_relmax05}
\small
\begin{tabular}{llccc}
\toprule
\textbf{Model A} & \textbf{Model B} & \textbf{$n$ tasks} & \textbf{$\rho$} & \textbf{$p$} \\
\midrule
\multicolumn{5}{l}{\textit{Within-family (OLMo-2)}} \\
OLMo2-1B & OLMo2-7B & 106 & 0.579 & $8.2 \times 10^{-11}$ \\
OLMo2-1B & OLMo2-13B & 105 & 0.433 & $3.9 \times 10^{-6}$ \\
OLMo2-7B & OLMo2-13B & 104 & 0.563 & $4.8 \times 10^{-10}$ \\
\midrule
\multicolumn{5}{l}{\textit{Within-family (Pythia)}} \\
Pythia-1.4B & Pythia-410M & 96 & 0.866 & $5.3 \times 10^{-30}$ \\
Pythia-1.4B & Pythia-12B & 98 & 0.828 & $8.1 \times 10^{-26}$ \\
Pythia-410M & Pythia-12B & 100 & 0.748 & $3.5 \times 10^{-19}$ \\
\midrule
\multicolumn{5}{l}{\textit{Within-family (LLM360)}} \\
Amber & Crystal & 102 & 0.489 & $1.8 \times 10^{-7}$ \\
\midrule
\multicolumn{5}{l}{\textit{Cross-family (OLMo-2 $\leftrightarrow$ OLMo-3)}} \\
OLMo2-1B & OLMo3-7B & 106 & 0.588 & $3.5 \times 10^{-11}$ \\
OLMo2-7B & OLMo3-7B & 105 & 0.702 & $7.1 \times 10^{-17}$ \\
OLMo2-13B & OLMo3-7B & 104 & 0.659 & $2.7 \times 10^{-14}$ \\
\midrule
\multicolumn{5}{l}{\textit{Cross-family (OLMo-2 $\leftrightarrow$ LLM360)}} \\
Amber & OLMo2-1B & 107 & 0.587 & $3.0 \times 10^{-11}$ \\
Amber & OLMo2-7B & 106 & 0.516 & $1.5 \times 10^{-8}$ \\
Amber & OLMo2-13B & 105 & 0.279 & $3.9 \times 10^{-3}$ \\
Crystal & OLMo2-1B & 102 & 0.743 & $3.9 \times 10^{-19}$ \\
Crystal & OLMo2-7B & 101 & 0.668 & $2.3 \times 10^{-14}$ \\
Crystal & OLMo2-13B & 100 & 0.571 & $5.5 \times 10^{-10}$ \\
\midrule
\multicolumn{5}{l}{\textit{Cross-family (OLMo-3 $\leftrightarrow$ LLM360)}} \\
Amber & OLMo3-7B & 106 & 0.513 & $1.9 \times 10^{-8}$ \\
Crystal & OLMo3-7B & 102 & 0.764 & $8.9 \times 10^{-21}$ \\
\midrule
\multicolumn{5}{l}{\textit{Cross-family (OLMo-2 $\leftrightarrow$ Pythia)}} \\
Pythia-1.4B & OLMo2-1B & 98 & 0.492 & $2.7 \times 10^{-7}$ \\
Pythia-1.4B & OLMo2-7B & 97 & 0.476 & $8.6 \times 10^{-7}$ \\
Pythia-1.4B & OLMo2-13B & 97 & 0.085 & $0.41$ \\
Pythia-410M & OLMo2-1B & 100 & 0.445 & $3.5 \times 10^{-6}$ \\
Pythia-410M & OLMo2-7B & 99 & 0.452 & $2.6 \times 10^{-6}$ \\
Pythia-410M & OLMo2-13B & 98 & 0.129 & $0.21$ \\
Pythia-12B & OLMo2-1B & 102 & 0.511 & $4.1 \times 10^{-8}$ \\
Pythia-12B & OLMo2-7B & 101 & 0.531 & $1.1 \times 10^{-8}$ \\
Pythia-12B & OLMo2-13B & 100 & 0.304 & $2.1 \times 10^{-3}$ \\
\midrule
\multicolumn{5}{l}{\textit{Cross-family (OLMo-3 $\leftrightarrow$ Pythia)}} \\
Pythia-1.4B & OLMo3-7B & 97 & 0.314 & $1.8 \times 10^{-3}$ \\
Pythia-410M & OLMo3-7B & 99 & 0.355 & $3.1 \times 10^{-4}$ \\
Pythia-12B & OLMo3-7B & 101 & 0.460 & $1.3 \times 10^{-6}$ \\
\midrule
\multicolumn{5}{l}{\textit{Cross-family (LLM360 $\leftrightarrow$ Pythia)}} \\
Amber & Pythia-1.4B & 98 & 0.763 & $7.3 \times 10^{-20}$ \\
Amber & Pythia-410M & 100 & 0.702 & $4.0 \times 10^{-16}$ \\
Amber & Pythia-12B & 102 & 0.753 & $6.9 \times 10^{-20}$ \\
Crystal & Pythia-1.4B & 93 & 0.381 & $1.6 \times 10^{-4}$ \\
Crystal & Pythia-410M & 96 & 0.341 & $6.6 \times 10^{-4}$ \\
Crystal & Pythia-12B & 97 & 0.409 & $3.1 \times 10^{-5}$ \\
\bottomrule
\end{tabular}
\end{table}

\begin{table}[h]
\centering
\caption{Pairwise Spearman rank correlations: Relative threshold ($\alpha = 0.8$, fraction of max performance). Mean $\rho = 0.500$.}
\label{tab:spearman_relmax08}
\small
\begin{tabular}{llccc}
\toprule
\textbf{Model A} & \textbf{Model B} & \textbf{$n$ tasks} & \textbf{$\rho$} & \textbf{$p$} \\
\midrule
\multicolumn{5}{l}{\textit{Within-family (OLMo-2)}} \\
OLMo2-1B & OLMo2-7B & 106 & 0.491 & $8.9 \times 10^{-8}$ \\
OLMo2-1B & OLMo2-13B & 105 & 0.359 & $1.7 \times 10^{-4}$ \\
OLMo2-7B & OLMo2-13B & 104 & 0.707 & $4.6 \times 10^{-17}$ \\
\midrule
\multicolumn{5}{l}{\textit{Within-family (Pythia)}} \\
Pythia-1.4B & Pythia-410M & 96 & 0.716 & $2.3 \times 10^{-16}$ \\
Pythia-1.4B & Pythia-12B & 98 & 0.547 & $5.5 \times 10^{-9}$ \\
Pythia-410M & Pythia-12B & 100 & 0.521 & $2.7 \times 10^{-8}$ \\
\midrule
\multicolumn{5}{l}{\textit{Within-family (LLM360)}} \\
Amber & Crystal & 102 & 0.632 & $1.0 \times 10^{-12}$ \\
\midrule
\multicolumn{5}{l}{\textit{Cross-family (OLMo-2 $\leftrightarrow$ OLMo-3)}} \\
OLMo2-1B & OLMo3-7B & 106 & 0.498 & $5.7 \times 10^{-8}$ \\
OLMo2-7B & OLMo3-7B & 105 & 0.773 & $4.8 \times 10^{-22}$ \\
OLMo2-13B & OLMo3-7B & 104 & 0.698 & $1.8 \times 10^{-16}$ \\
\midrule
\multicolumn{5}{l}{\textit{Cross-family (OLMo-2 $\leftrightarrow$ LLM360)}} \\
Amber & OLMo2-1B & 107 & 0.556 & $5.3 \times 10^{-10}$ \\
Amber & OLMo2-7B & 106 & 0.612 & $3.0 \times 10^{-12}$ \\
Amber & OLMo2-13B & 105 & 0.544 & $2.1 \times 10^{-9}$ \\
Crystal & OLMo2-1B & 102 & 0.634 & $8.8 \times 10^{-13}$ \\
Crystal & OLMo2-7B & 101 & 0.714 & $5.0 \times 10^{-17}$ \\
Crystal & OLMo2-13B & 100 & 0.603 & $3.1 \times 10^{-11}$ \\
\midrule
\multicolumn{5}{l}{\textit{Cross-family (OLMo-3 $\leftrightarrow$ LLM360)}} \\
Amber & OLMo3-7B & 106 & 0.590 & $2.8 \times 10^{-11}$ \\
Crystal & OLMo3-7B & 102 & 0.674 & $8.6 \times 10^{-15}$ \\
\midrule
\multicolumn{5}{l}{\textit{Cross-family (OLMo-2 $\leftrightarrow$ Pythia)}} \\
Pythia-1.4B & OLMo2-1B & 98 & 0.580 & $4.0 \times 10^{-10}$ \\
Pythia-1.4B & OLMo2-7B & 97 & 0.286 & $4.5 \times 10^{-3}$ \\
Pythia-1.4B & OLMo2-13B & 97 & 0.077 & $0.45$ \\
Pythia-410M & OLMo2-1B & 100 & 0.502 & $1.0 \times 10^{-7}$ \\
Pythia-410M & OLMo2-7B & 99 & 0.309 & $1.9 \times 10^{-3}$ \\
Pythia-410M & OLMo2-13B & 98 & 0.159 & $0.12$ \\
Pythia-12B & OLMo2-1B & 102 & 0.523 & $1.7 \times 10^{-8}$ \\
Pythia-12B & OLMo2-7B & 101 & 0.527 & $1.5 \times 10^{-8}$ \\
Pythia-12B & OLMo2-13B & 100 & 0.383 & $8.3 \times 10^{-5}$ \\
\midrule
\multicolumn{5}{l}{\textit{Cross-family (OLMo-3 $\leftrightarrow$ Pythia)}} \\
Pythia-1.4B & OLMo3-7B & 97 & 0.177 & $0.08$ \\
Pythia-410M & OLMo3-7B & 99 & 0.298 & $2.8 \times 10^{-3}$ \\
Pythia-12B & OLMo3-7B & 101 & 0.436 & $5.1 \times 10^{-6}$ \\
\midrule
\multicolumn{5}{l}{\textit{Cross-family (LLM360 $\leftrightarrow$ Pythia)}} \\
Amber & Pythia-1.4B & 98 & 0.517 & $4.9 \times 10^{-8}$ \\
Amber & Pythia-410M & 100 & 0.437 & $5.5 \times 10^{-6}$ \\
Amber & Pythia-12B & 102 & 0.622 & $3.0 \times 10^{-12}$ \\
Crystal & Pythia-1.4B & 93 & 0.399 & $7.3 \times 10^{-5}$ \\
Crystal & Pythia-410M & 96 & 0.376 & $1.6 \times 10^{-4}$ \\
Crystal & Pythia-12B & 97 & 0.512 & $8.5 \times 10^{-8}$ \\
\bottomrule
\end{tabular}
\end{table}

\newpage


\section{Function vector hyperparameters}
\label{appendix:fv_hyperparams}

\autoref{tab:fv-hyperparams} shows the hyperparameters selected for each model's task representation. Hyperparameters (representation type -- between a fixed set of heads and a full residual stream, layers, and number of heads) were chosen via a three-criterion search over candidate configurations. A fixed calibration set of elemental and composite tasks was used, and three criteria were considered: (1) \textit{within-task consistency}, measured as split-half cosine similarity between FVs extracted from random partitions of correct examples; (2) \textit{inter-task discriminability}, the ratio of within-task to between-task cosine similarity; and (3) \textit{compositional structure}, measured as the cosine similarity between each compositional task's FV and its least-squares reconstruction from component FVs. Final selection used a rank-sum policy over these three criteria, with ties broken by raw metric values. Note that in all cases, only correct examples were used when constructing the final function vectors.

\begin{table}[t]
\centering
\small
\begin{tabular}{llrrrl}
\toprule
\textbf{Model} & \textbf{Representation} & \textbf{Layer} & \textbf{\textit{k} heads} & \textbf{$\sigma$} & \textbf{$\lambda$} \\
\midrule
\multicolumn{6}{l}{\textit{LLM360}} \\[0.2em]
\texttt{amber}        & hidden states &  21 & ---  & 6.02568 & 0.0001 \\
\addlinespace[0.4em]
\texttt{crystal}      & hidden states &   8 & ---  & 6.25822 & 0.0001 \\
\addlinespace[1em]
\multicolumn{6}{l}{\textit{Pythia}} \\[0.2em]
\texttt{pythia\_410m} & hidden states &   3 & ---  & 0.33991 & 0.001  \\
\addlinespace[0.4em]
\texttt{pythia\_1.4b} & hidden states &  12 & ---  & 5.93639 & 0.001  \\
\addlinespace[0.4em]
\texttt{pythia\_12b}  & hidden states &   9 & ---  & 4.02777 & 0.0001 \\
\addlinespace[1em]
\multicolumn{6}{l}{\textit{OLMo-2}} \\[0.2em]
\texttt{olmo2\_1b}    & hidden states &   8 & ---  & 3.46810 & 0.0001 \\
\addlinespace[0.4em]
\texttt{olmo2\_7b}    & hidden states &  16 & ---  & 1.05641 & 0.005  \\
\addlinespace[0.4em]
\texttt{olmo2\_13b}   & cie\_heads       &  10 &  15  & 0.96582 & 0.005  \\
\addlinespace[1em]
\multicolumn{6}{l}{\textit{OLMo-3}} \\[0.2em]
\texttt{olmo3\_7b}    & hidden states &  16 & ---  & 4.37314 & 0.001  \\
\bottomrule
\end{tabular}
\caption{Function vector hyperparameters selected per model. The full residual stream was chosen as the representation for all models besides OLMo2-13B, in which the top 10 heads by causal indirect effect in layer 10 were chosen. $\sigma$ and $\lambda$ are the parameters used in ridge regression.}
\label{tab:fv-hyperparams}
\end{table}

\section{All Held-out trajectory predictions}
\label{appendix:all_held_out_preds}

\begin{figure}
    \centering
    \includegraphics[width=0.95\linewidth]{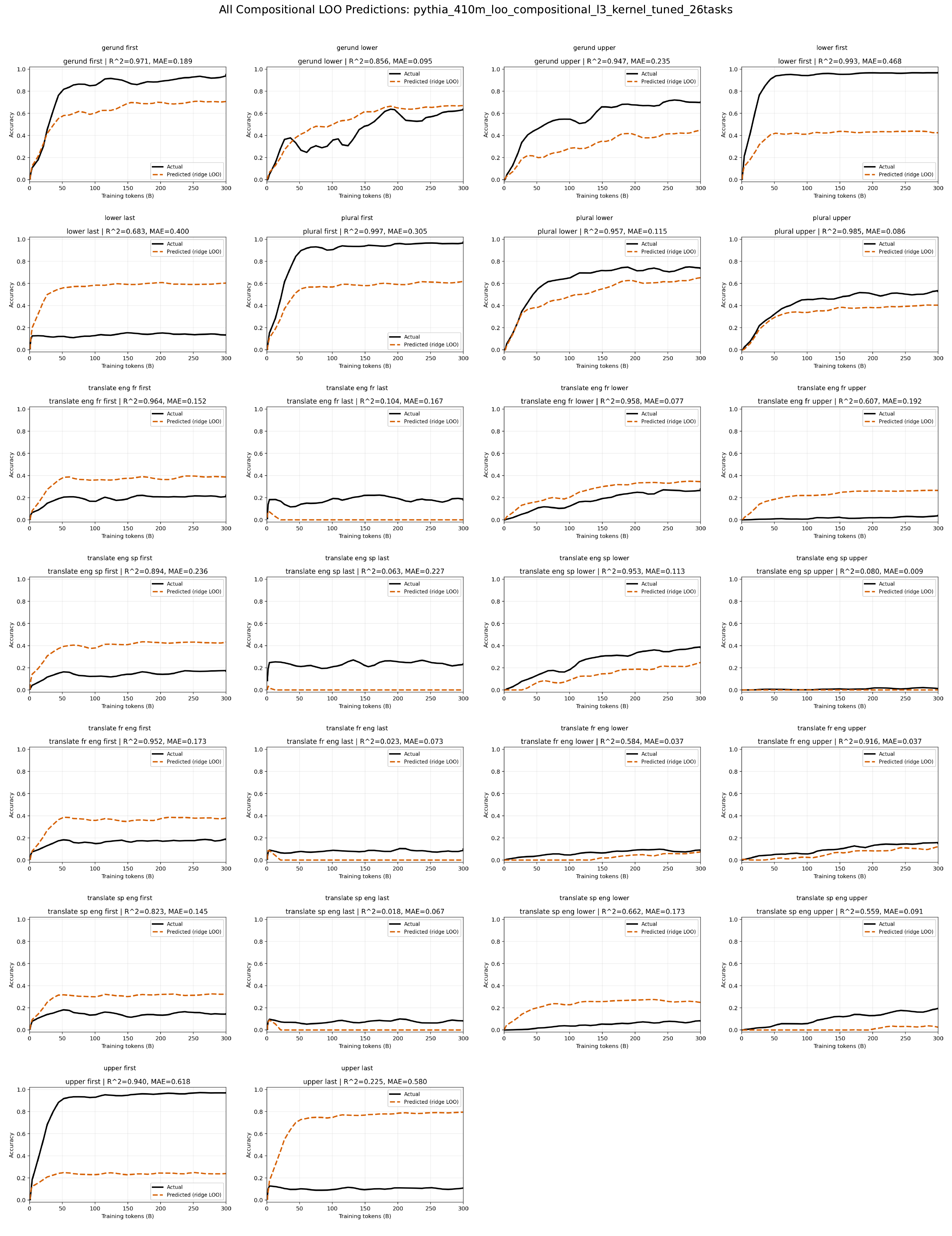}
    \caption{Predictions of held-out compositional tasks for Pythia-410M.}
    \label{fig:held_out_pred_pythia_410m}
\end{figure}

\begin{figure}
    \centering
    \includegraphics[width=0.95\linewidth]{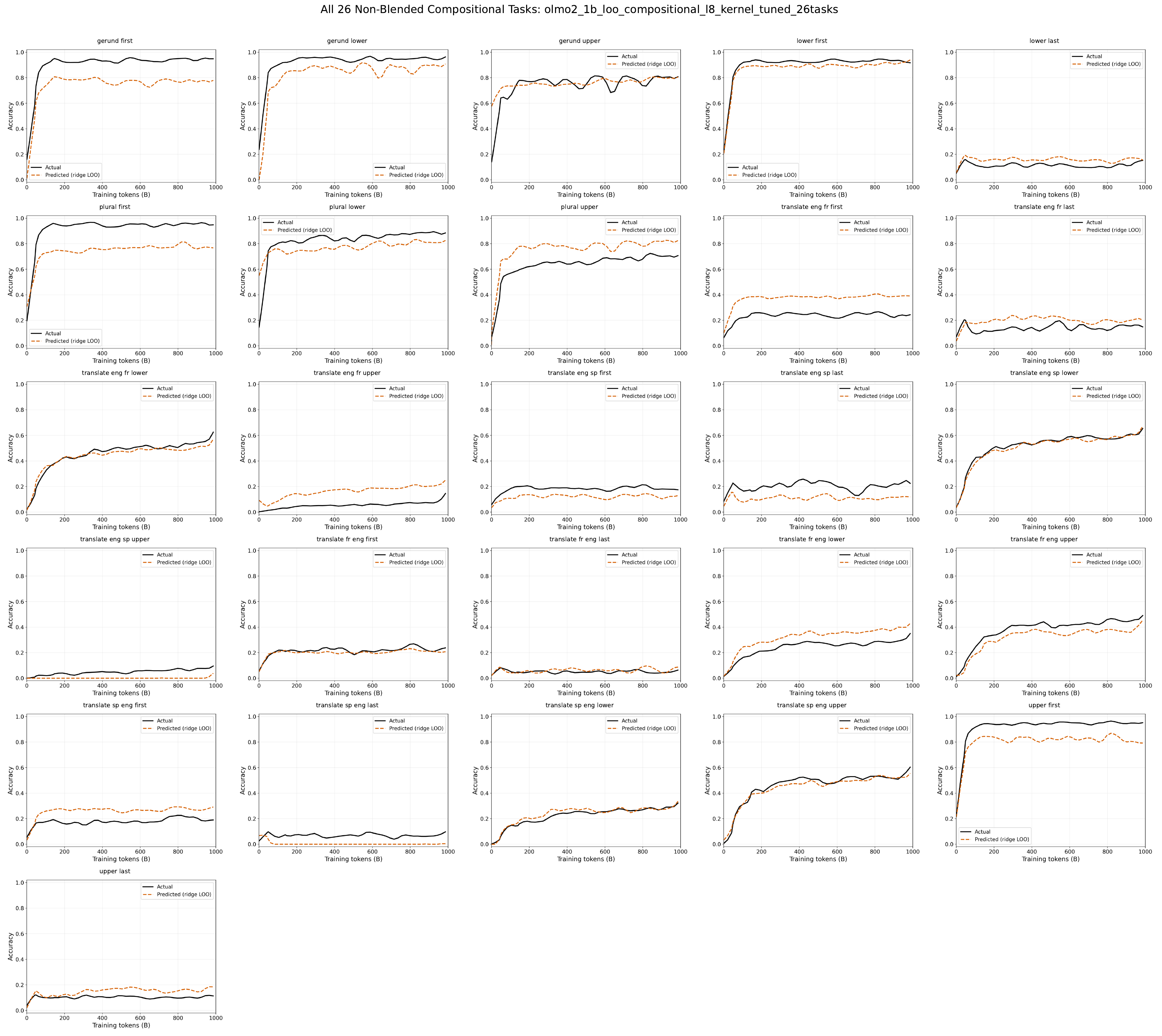}
    \caption{Predictions of held-out compositional tasks for OLMo2-1B.}
    \label{fig:held_out_pred_olmo2_1b}
\end{figure}

\begin{figure}
    \centering
    \includegraphics[width=0.95\linewidth]{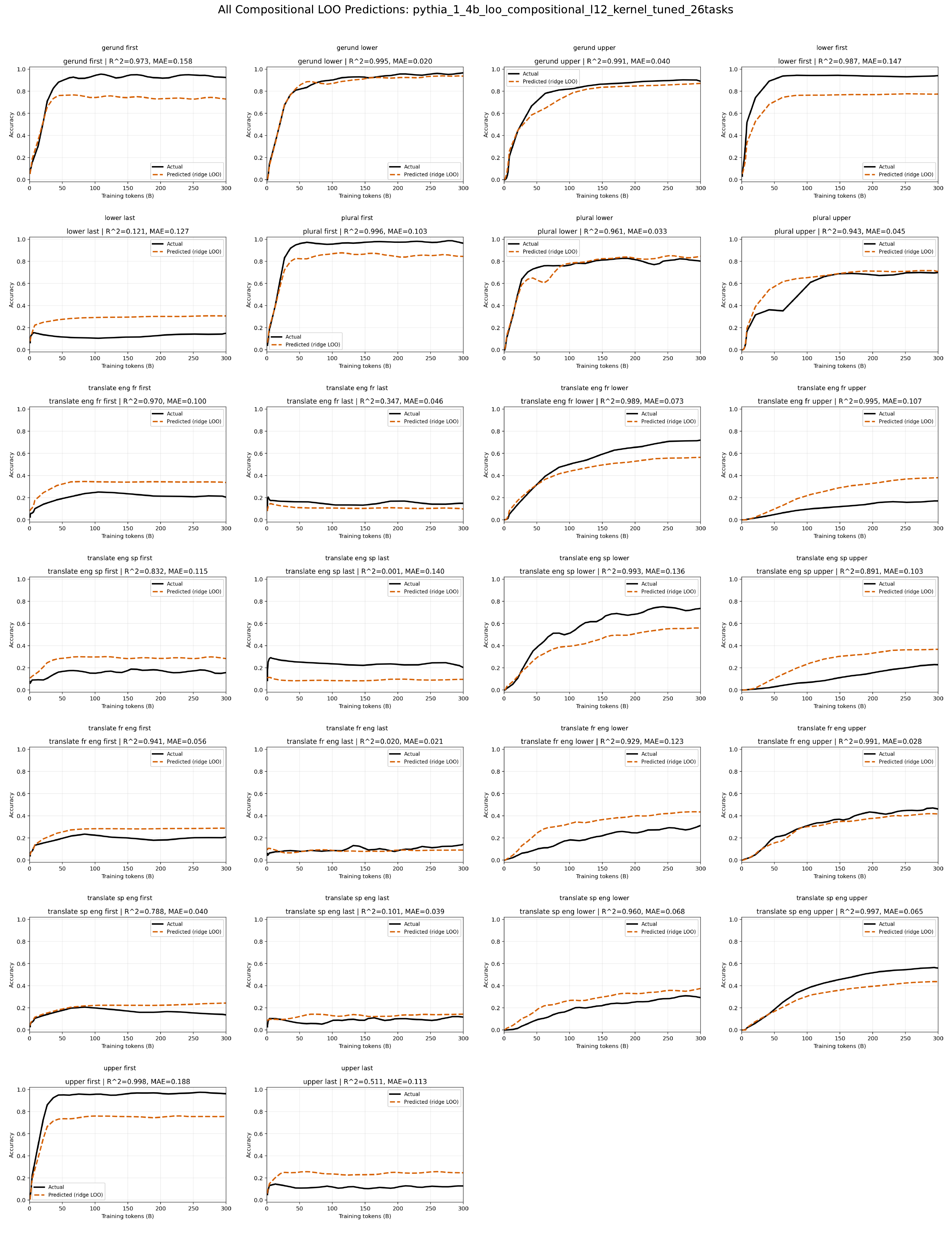}
    \caption{Predictions of held-out compositional tasks for Pythia-1.4B.}
    \label{fig:held_out_pred_pythia_1_4b}
\end{figure}

\begin{figure}
    \centering
    \includegraphics[width=0.95\linewidth]{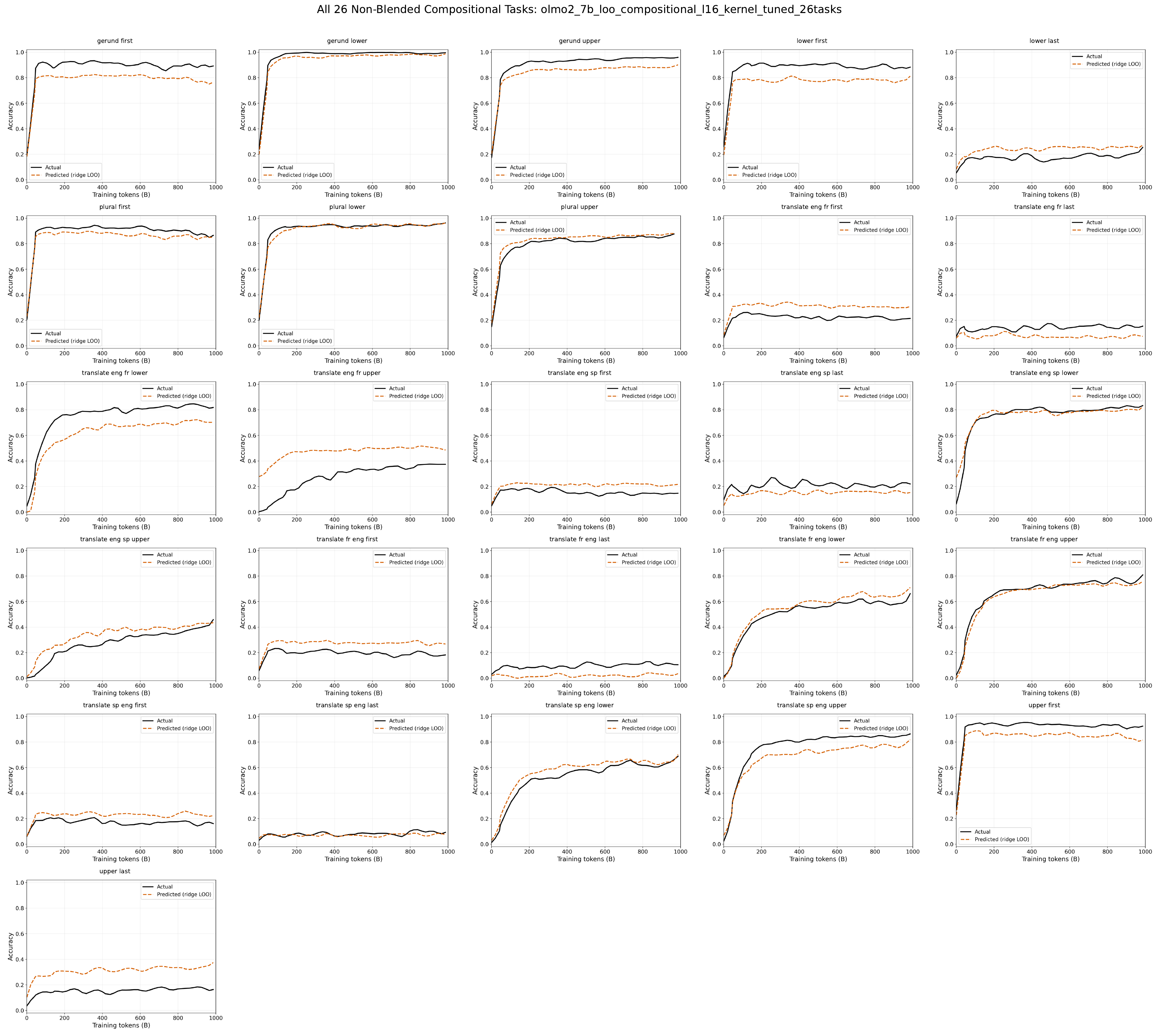}
    \caption{Predictions of held-out compositional tasks for OLMo2-7B.}
    \label{fig:held_out_pred_olmo2_7b}
\end{figure}

\begin{figure}
    \centering
    \includegraphics[width=0.95\linewidth]{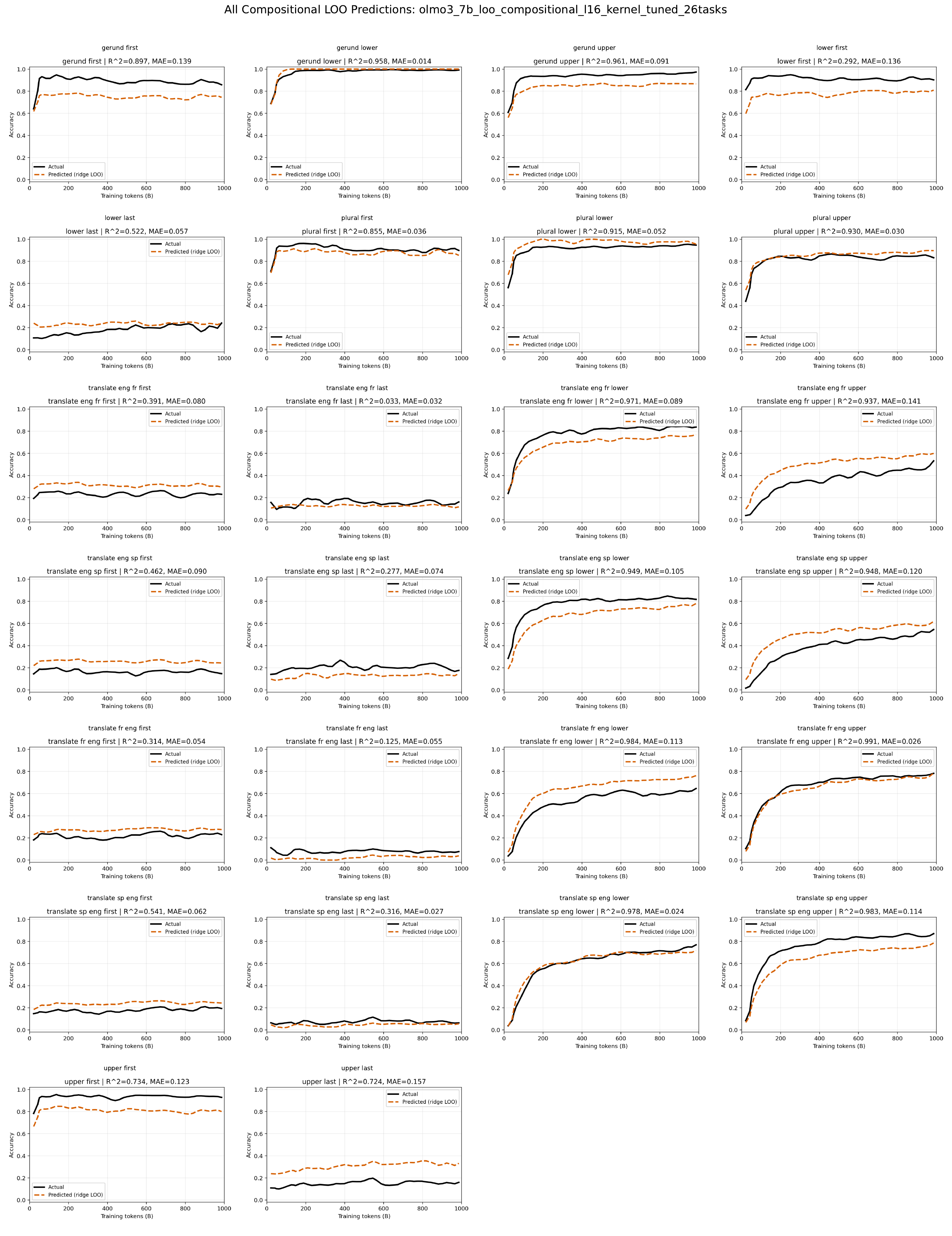}
    \caption{Predictions of held-out compositional tasks for OLMo3-7B.}
    \label{fig:held_out_pred_olmo3_7b}
\end{figure}

\begin{figure}
    \centering
    \includegraphics[width=0.95\linewidth]{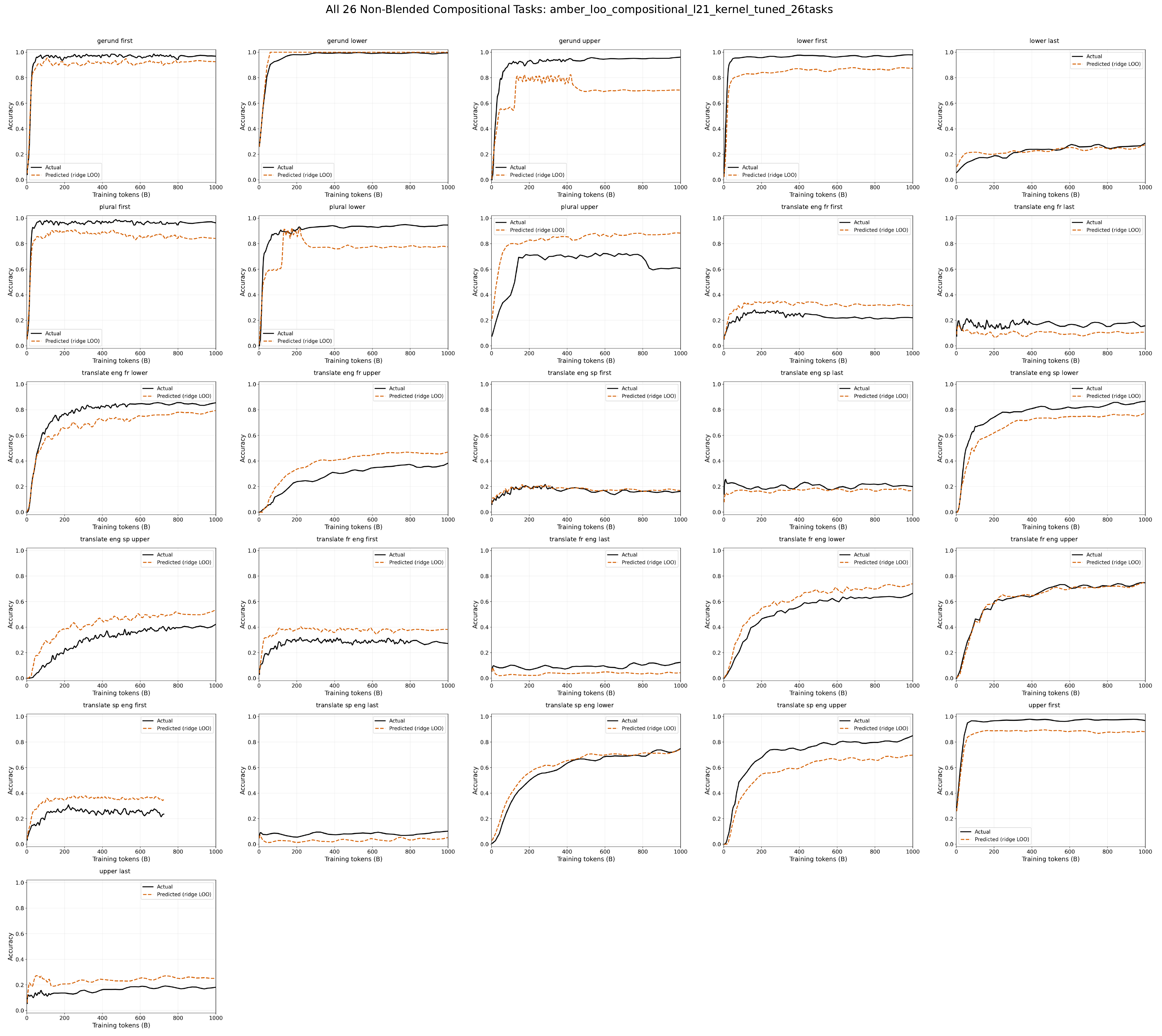}
    \caption{Predictions of held-out compositional tasks for Amber.}
    \label{fig:held_out_pred_amber}
\end{figure}

\begin{figure}
    \centering
    \includegraphics[width=0.95\linewidth]{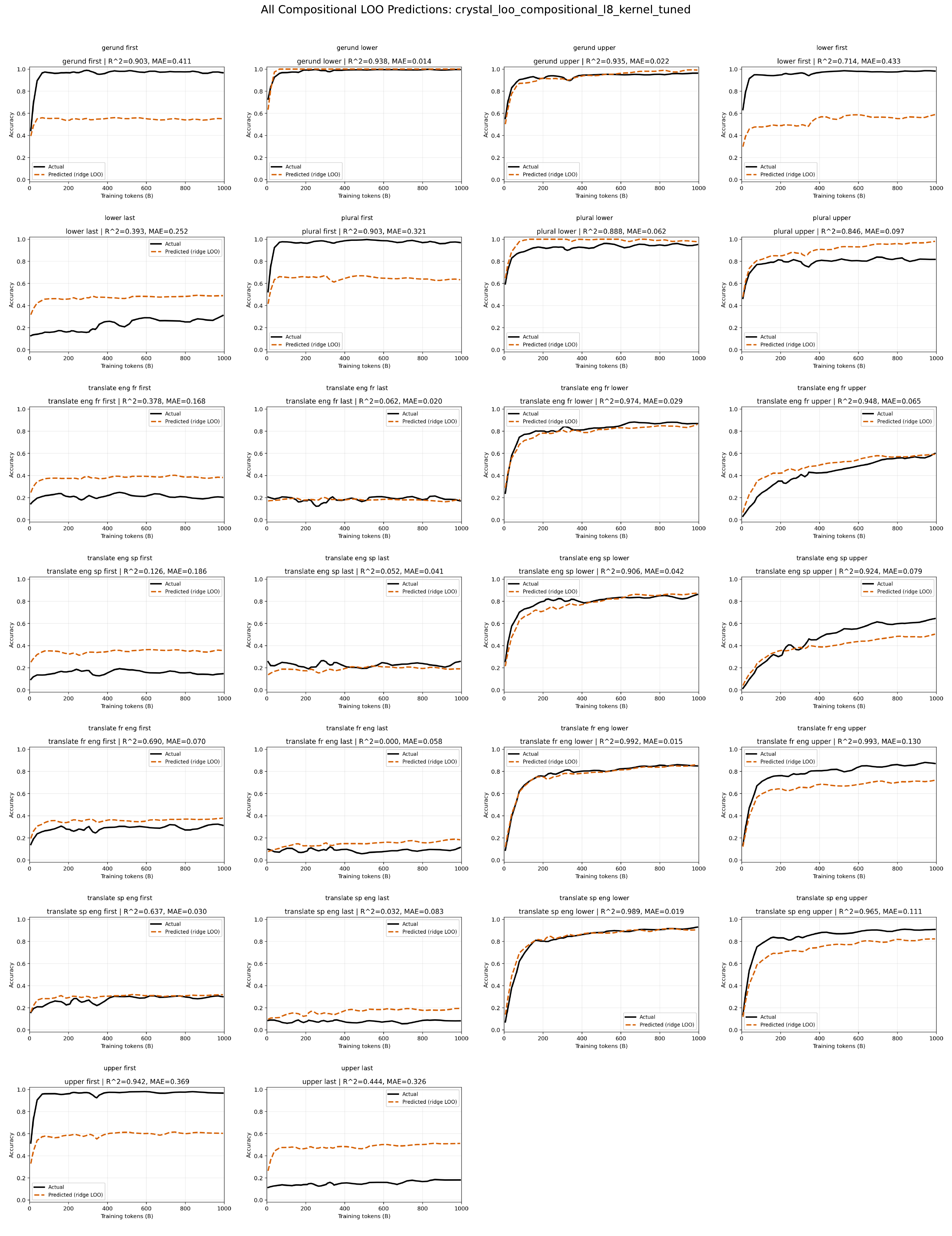}
    \caption{Predictions of held-out compositional tasks for CrystalCoder.}
    \label{fig:held_out_pred_crystal}
\end{figure}

\begin{figure}
    \centering
    \includegraphics[width=0.95\linewidth]{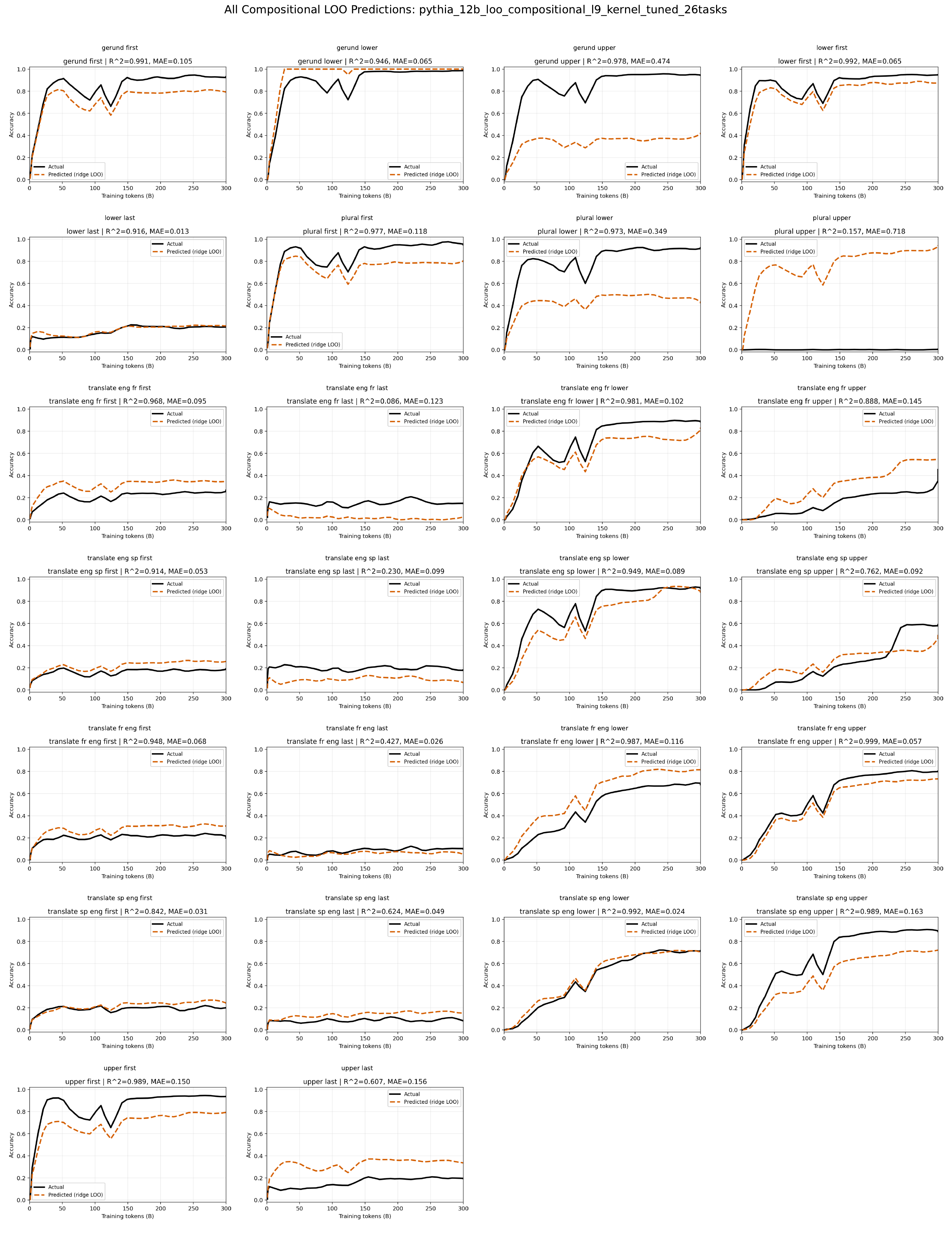}
    \caption{Predictions of held-out compositional tasks for Pythia-12B.}
    \label{fig:held_out_pred_pythia_12b}
\end{figure}

\begin{figure}
    \centering
    \includegraphics[width=0.95\linewidth]{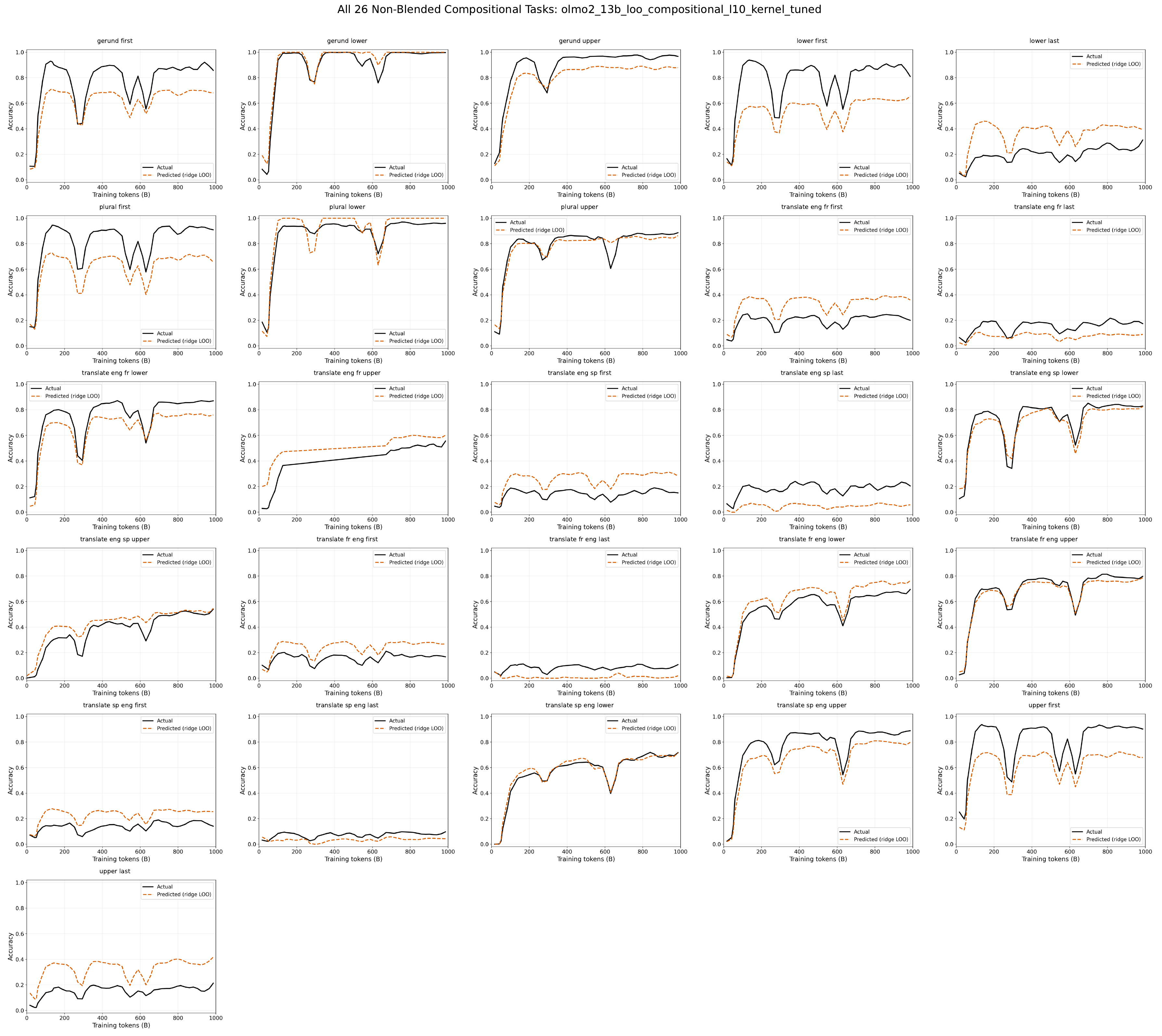}
    \caption{Predictions of held-out compositional tasks for OLMo2-13B.}
    \label{fig:held_out_pred_olmo2_13b}
\end{figure}

Figures \ref{fig:held_out_pred_pythia_410m} -- \ref{fig:held_out_pred_olmo2_13b} show the results of the leave-one-out prediction setup for each compositional task.

\end{document}